\documentclass[12pt,twoside,reqno]{amsart}
\usepackage[colorlinks=true,citecolor=blue]{hyperref}
\usepackage{mathptmx, amsmath, amssymb, amsfonts, amsthm, mathptmx, enumerate, color, colortbl, threeparttable}
\usepackage{graphicx}
\usepackage{epstopdf}
\usepackage{subfigure}

\theoremstyle{definition}
\newtheorem{definition}{Definition}[section]

\numberwithin{equation}{section}

\definecolor{DarkGray}{gray}{0.7}
\definecolor{LightGray}{gray}{0.9}

\begin{document}
\setcounter{page}{1}

\vspace*{1.0cm}
\title[]
{The Analysis from Nonlinear Distance Metric to Kernel-based Drug Prescription Prediction System}
\author[]{Der-Chen Chang$^{1,2}$, Ophir Frieder$^1$, Chi-Feng Hung$^3$, Hao-Ren Yao$^{1*}$}
\maketitle
\vspace*{-0.6cm}

\begin{center}
{\footnotesize {\it
$^1$Department of Computer Science, Georgetown University, Washington, D.C., USA \\
$^2$Department of Mathematics $\&$ Statistics, Georgetown University, Washington, D.C., USA \\
$^3$School of Medicine, Fu Jen Catholic University, New Taipei City, Taiwan
}}\end{center}

\vskip 4mm {\small\noindent {\bf Abstract.}
The distance metric and its nonlinear variant play a substantial role in machine learning, particularly yoso in building kernel functions. Often, the Euclidean distance with a radial basis function (RBF) is used to construct a RBF kernel for nonlinear classification. However, domain implications periodically constrain the distance metrics.  Specifically, within the domain of drug efficacy prediction, distance measures must account for time that varies based on disease duration, short to chronic. 

Recently, a distance-derived graph kernel approach was commercially licensed for drug prescription efficacy prediction.  The analysis of the distance functions used therein, namely the Euclidean and cosine distance measures and their respective derived graph kernels, is provided. Theoretically, we provide a formulation of our efforts and demonstrate how both the Euclidean and cosine distance induce space and discuss the difference from geometric perspectives.  

The aforementioned approach is likewise empirically evaluated using a million-plus patient subset of a life-spanning, real-world, electronic health record database. Diseases are characterized as either short in duration or chronic and either common, hence balanced data, or relatively rare, hence imbalanced.  Empirically, the system accurately predicted the efficacy of prescriptions for both balanced and imbalanced and short-term and chronic diseases, with at least one of the measures used being statistically significantly superior to conventional prediction methods.  Succinctly, for short-term, balanced diseases, the Euclidean and cosine measures were generally statistically equivalent.  For short-term, imbalanced diseases however, the Euclidean measure was superior to the cosine measure, at times and not infrequently, statistically significantly so.  For chronic, balanced diseases, Euclidean was slightly superior to the cosine measure, but they were statistically equivalent.  In contrast, for chronic, imbalanced diseases, the cosine measure was consistently statistically significantly superior to the Euclidean measure.  These findings indicate the need for both measures depending on the use case.  Our empirical findings match our theoretical underpinnings.  

 }

\renewcommand{\thefootnote}{}
\footnotetext{ $^*$Corresponding author.
\par
E-mail addresses: hao-ren@ir.cs.georgetown.edu (Hao-Ren Yao)
\par
Received XXX; Accepted February XXX. }

\section{Introduction}
\label{introduction}
Distance metrics and their nonlinear variants play a fundamental role in machine learning tasks. They measure the degree of linear or nonlinear similarity between objects, grouping similar objects into $k$ different groups (e.g., $k$-means clustering) or assigning class labels based on their nearest neighbors (e.g, $k$-nearest neighbor classification) ~\cite{k-means}. There are multiple choices of distance functions to solve specific problems. Generally, Euclidean distance is used for the majority of classification problems, whereas, cosine distance is more suitable for document classification ~\cite{text-mining-survey, ruocco}. Distance metrics can also be part of kernel function design ~\cite{JNCA}. In ~\cite{bcb-2019}, the authors use Euclidean distance as part of a graph kernel to capture time difference similarity between patients. In ~\cite{dist-sub-kernel}, the authors demonstrate the positive definiteness for a distance substitution kernel. Usually, the kernel function $k$ for two data objects $x$ and $y$ and user defined parameter $\alpha$ is defined as:

\begin{equation*}
k_t(x, y) \,=\,C_t e^{-\alpha\|x-y\|},
\end{equation*}

\noindent
where $\|x-y\|$ is the Euclidean distance between $x$ and $y$, and $C_t$ is a constant depending on time $t$.

Distance metric learning, on the other hand, tries to find the optimal distance metric or embedding space given a task specific objective for a set of data objects ~\cite{metric-learning-survey}. Under this scope, semantically similar objects are encouraged to be closer to each other and further apart for non-similar objects. The prevalence of deep learning has largely populated the development of deep metric learning, where nonlinear observation can be captured, specifically in unsupervised representation learning for images (e.g., computer vision ~\cite{computer-vision-related, deep-metric-learning}). It also motivated the deep kernel learning ~\cite{deep-kl1, deep-kl2}. In ~\cite{bcb-2020}, a deep metric learning based graph kernel is proposed to solve the outcome prediction problem in complex chronic disease treatment planning, where cosine distance was shown superior to the Euclidean distance measure.

In practice, both distance metric and kernel function are used to solve complex real-world problems, especially in medicine. A problem of interest is drug prescription efficacy prediction ~\cite{drug-1, drug-2}. Accurate predictive models for drug prescription improve healthcare. They can further reduce medication errors and identify possible drug prescription pathways to pursue for clinical personnel. In ~\cite{bhi-2019}, a framework is proposed to predict success and failure outcomes of a given drug prescription for antibiotic treatment-based disease. The approach is further extended to overcome biased data distribution ~\cite{bcb-2019} and chronic disease treatment plan ~\cite{bcb-2020}. Moreover, Euclidean and cosine distance measures exhibit different performance behaviors. In ~\cite{bcb-2020}, cosine distance is found superior to the Euclidean measure under highly data imbalanced chronic disease, however, Euclidean distance is used for short-term disease in ~\cite{bcb-2019}. To further investigate such differences, we establish a unified framework, which integrates two model structures proposed in the aforementioned works, to conduct rigorous empirical evaluation on all diseases investigated in previous studies ~\cite{bcb-2019, bcb-2020}, in addition to a theoretical discussion.  The aforementioned prescription efficacy prediction approaches are now under commercial licensing.

Our contributions are as follows:

\begin{itemize}
    \item We propose a scalable unified framework for prescription efficacy prediction.
    \item We evaluate performance using 10-fold cross validation on large-scale, real-world, electronic health record dataset that includes common and rare, short and chronic illnesses.
    \item We investigate the difference between Euclidean and cosine distance on learned embedding space.
    \item We provide a theoretical explanation from geometry perspective, generalizing Euclidean and cosine distance.
\end{itemize}

\section{Preliminaries}
Our previous efforts ~\cite{bhi-2019, bcb-2019, bcb-2020} present a graph kernel-based system for outcome prediction of drug prescription, particularly the success or failure treatment, on short-term and chronic diseases. In ~\cite{bcb-2019}, a Multiple Graph Kernel Fusion (MGKF) is proposed to overcome noise effect on short-term disease. A deep graph kernel learning approach, e.g., Cross-Global Attention Graph Kernel Network (Cross-Global), is proposed in ~\cite{bcb-2020} to handle long-term chronic disease. In short, we initially determine success and failure patients for the target disease treatment as training data within a user-defined time quantum, where a set of medical events are extracted between this time period. Then, we construct a patient graph, a graphical representation of the patient EHRs, given the extracted medical events. Finally, we perform binary graph classification as prediction through a graph kernel and a kernel-based classifier. We detail each part of the prediction framework in this section.

\subsection{Outcome Selection}
We define a failure drug prescription or treatment plan for a disease diagnosis if certain events occur within a predefined time period, otherwise a success. In short-duration diseases, we observe the similar or identical type of disease diagnosis, while in chronic disease, the observed target will be severe complication defined via medication guideline. We name a disease as short-term if it only considers a single medication with immediate outcome observation and recent medical history (e.g., 2 months prior to the diagnostic). For a chronic disease, we consider a multiple medication treatment plan with long-term outcome observation and medical history (e.g., 10 years prior to the diagnostic). We refer readers to MGKF ~\cite{bcb-2019} and Cross-Global ~\cite{bcb-2020} for greater detail.

\subsection{Patient Graph}
A subset of patient's EHRs is formulated as directly acyclic graph where a node represents each medical event and an edge with time difference (e.g., days) used as a weight connects two consecutive medical events. Patient demographic information, such as gender and age, is included by connecting to the first medical event with age as an edge weight$^1$\footnote{$^1$To simplify model assumption, we only use gender and age as demographic information.}. We define a patient graph here as in ~\cite{bcb-2019} and ~\cite{bcb-2020}:

\begin{definition}[\textit{Patient Graph}]
Given $n$ medical events, set $M = \{({m_1, t_1}),\ldots,({m_n, t_1})\}$ represents a patient's EHR with $m_i$ denoting a medical event such as diagnosis, and $t_i$ denoting the time for $m_i$. The patient graph $P_g=(V,E)$ of events $M$ is a weighted directed acyclic graph with its vertices $V$ containing all events $m_i \in M$ and edges $E$ containing all pairs of  consecutive events $(m_i, m_j)$. The edge weight from node i to node j is defined as $W_{ij} = t_j - t_i$ which defines the time interval between $m_i,m_j$.
\end{definition}

\subsection{Graph Kernel}
For kernel-based binary graph classification, building a pairwise kernel matrix between patient graphs is the first step. The graph kernel computes the similarity between pairs of graph. It is also a positive definite or semidefinite kernel defined for graphs, which performs an inner product implicitly by mapping data point from an input space to the Hilbert space. It can also be treated as a similarity measurement between two data objects (e.g, graphs).  We point readers to these articles ~\cite{gk-survey} for a more in-depth graph kernel discussion and ~\cite{JNCA} for a better understanding and design principle of graph kernel and its associated feature maps. In ~\cite{bhi-2019, bcb-2019}, several graph kernels are proposed to solve the drug prescription outcome prediction problem as patient graph classification. Please refer ~\cite{bhi-2019, bcb-2019} for more in-depth descriptions on kernel definitions.

\subsection{Prediction Framework}
We then formulate a binary graph classification problem on the resulting patient graph by using a kernelized Support Vector Machine (K-SVM) ~\cite{bhi-2019}, Multiple Graph Kernel Fusion (MGKF) ~\cite{bcb-2019}, and Cross-Global Attention Graph Kernel Network (Cross-Global) ~\cite{bcb-2020}.

As mentioned in ~\ref{introduction}, cosine distance is superior to its Euclidean counterpart under highly imbalanced chronic disease (Cross-Global), while both the Euclidean and cosine distance measures achieve high prediction performance under short-term diseases (MGKF). We now question how they relate to each other. It is not suitable to directly compare Cross-Global and MGKF since the following:

\begin{itemize}
    \item Different datasets$^2$~\footnote{$^2$Different database provider.} are used in MGKF and Cross-Global.  
    \item Different model structures and optimization perspectives. In MGKF, optimization aims at generating optimal kernel fusion, while it turns out to find optimal graph embedding under Cross-Global.
    \item Different data balance and imbalance ratio between short-term and chronic disease.
\end{itemize}

To fairly compare MGKF and Cross-Global with Euclidean and cosine distance under short-term and chronic disease, a unified framework is required. Here, we extend and generalize previous efforts to differentiate the behavior of Euclidean and cosine distance, in addition to the theoretical discussion. A unified framework for graph-kernel based drug prescription outcome prediction is presented to conduct a rigorous empirical evaluation on all diseases in previous works on a very large-scale real-world EHRs.

\section{Discussion from the Geometric Point of View}
\subsection{Riemannian and SubRiemannian geometries} 
To discuss differences between Euclidean and cosine distance, we first establish some mathematical properties with those distance under geometry point of view. In fact, we may consider this problem in a more general setting. 

Let ${\mathcal X}=\{X_1,\ldots, X_m\}$ be $m$ linearly independent vector fields on an $n$-dimensional real manifold $\mathcal M_n$ with $m\le n$ of the tangent bundle $T{\mathcal M}_n$.
To find a good kernel function to describe the diffusion (energy flow) between two points in $\mathcal M_n$, we need to solve the 
heat equation associate to the the sum of square of $X_j$'s:
$$
\Delta_{\mathcal X}\,=\, \frac 12\sum_{j=1}^m X_j^2.
$$
When $m=n$, the operator $\Delta_{\mathcal X}$ is elliptic. Assume that  
$$
X_j\perp X_k,\qquad 1\le j,k\le n,\qquad j\not= k,\qquad {\text{and}}\qquad \|X_j\|\,=\, 1.
$$
In this case, we have a natural volume element which yields the adjoint vector fields $X_j^\ast$ for $1\le j\le m$. 
More precisely, for $\phi,\,\psi\in C^\infty_0(\mathcal M_n)$, 
$$
\int_{\mathcal M_n} \big(X_j \phi \big) \psi\,=\, \int_{\mathcal M_n} \phi \big(X_j^\ast  \psi\big),
$$
whence 
$-\frac 12\sum_{j=1}^m X_j^\ast X_j$ 
is the classical Laplace-Beltrami operator whose second order part agrees with the operator $\Delta_{\mathcal X}$. This suggests us that we may use 
the given differential operator  $\Delta_{\mathcal X}$ to introduce a geometry on $\mathcal M_n$ which may help us to solve $\Delta_{\mathcal X}$ and hence the heat equation. 
Hence, for $n\ge 2$, the solving kernel for the heat operator $\frac{\partial}{\partial t}-\Delta_{\mathcal X}$ takes the form
$$
{\mathcal P}_t(x,x_0)=\frac{1}{(2\pi t)^{\frac{n}{2}}}e^{-\frac{d^2(x,x_0)}{2t}}\big(a_0+a_1t+a_2t^2+\cdots\big).
$$
The $a_j$'s are functions of $x$ and $x_0$. Here $d(x,x_0)$ represents the induced Riemannian distance between the points $x_0$ and $x$ in $\mathcal M_n$. 
Moreover, $+\cdots$ stands for negligible error.  Furthermore, 
$$
\frac{\partial}{\partial t}\Big(\frac{d^2(x,x_0)}{2t}\Big)+\frac 12\sum_{j=1}^n\Big(X_j\frac{d^2(x,x_0)}{2}\Big)^2\,=\,
\frac{\partial}{\partial t}\Big(\frac{d^2(x,x_0)}{2t}\Big)+\frac 12H\Big(X_1(\frac{d^2}{2t}),\ldots, X_n(\frac{d^2}{2t})\Big)
\,=\, 0,
$$
{\it i.e.,} $\frac{d^2(x,x_0)}{2t}$ is a solution of the {\it Hamilton-Jacobi equation}. Here 
$H$ is the Hamiltonian function associated with $\Delta_{\mathcal X}$.

The simplest example will be the Euclidean distance $d(x,y)=\|x-y\|$ and the kernel ${\mathcal P}_t(x,x_0)$ is the Gaussian (see \cite{CC2}). In this paper, we are going to use another non-trivial example of Riemannian metric.  
Given a large sample space, we first embed those sample s in an $n$-dimensional sphere $\mathcal M_n=\mathbb S^{n-1}$. Given two points $x_1$ and $x_2$ in $\mathcal M_n$, we define the ``distance" $d(x_1,x_2)=1-\cos\big(\frac{\langle x_1,x_2\rangle}{\|x_1\|\cdot \|x_2\|}\big)$. Here $\| x\|$ is the Euclidean distance between the point $x$ and the origin. This is so-called {\it cosine metric}.  In other words, $x_1$ and $x_2$ are on the same sphere. Hence, $x_1$ and $x_2$ are located on a "big circle" which is determined by the center $x_0$ of the sphere and these two points. Instead of measure the arc-length (which maybe huge), we consider the angle $\theta$ between $\vec{x_0x_1}$ and $\vec{x_0x_2}$. This metric provides better estimates for the kernel in applications of drug prescription prediction system for long term disease.  

When $m<n$, the operator is non-elliptic. In this case, the subspace $\mathcal D\,=\, {\mbox{span}}\{X_1,\ldots, X_m\}$ 
is called the ``horizontal subspace" of $T{\mathcal M}_n$, and the vectors $X\in \mathcal D_p$ are called horizontal vectors at $p$.
Sometimes, we call the distribution $\mathcal D$ the {\it horizontal distribution}. 
The sections in the horizontal bundle $(\mathcal D,\mathcal M_n)$ are called horizontal vector fields. They are smooth assignments $\mathcal M_n\,\ni\, p\,\to\, X_p\in\mathcal D_p$. 
The set of the horizontal vector fields on $\mathcal M_n$ will be denoted by $\Gamma(\mathcal D)$. If $U$ is an open subset of $\mathcal M_n$, the set of horizontal vector fields on $U$ will be denoted by $\Gamma(\mathcal D, U)$. 
We call the complement of $\Gamma(\mathcal D, U)$ the ``{\it missing direction}" at $U$. 

Now we encounter new problems since $\mathcal D \not= T{\mathcal M}_n$, and we cannot find arc-length in general. 
We overcome this difficulty by assuming bracket generating property: ``the horizontal vector fields $\mathcal X$ and their brackets span 
$T{\mathcal M}_n$", then Chow's theorem \cite{CHOW} to conclude that given any two points
$A,B\in {\mathcal M}_n$, there is a piecewise $C^1$ horizontal curve $\gamma:[0,1]\, \rightarrow\, {\mathcal M}_n$ such that 
\[ 
\gamma(0)=A,\quad \gamma(1)=B,
\] 
and 
\[
\dot\gamma(s)=\sum_{k=1}^m a_k(s)X_k. 
\] 
This yields a distance, and therefore a geometry, which we shall call {\it subRiemannian geometry}. SubRiemannian geometry was first discussed in the field of thermodynamics around 1800s. Carnot discovered  the principle of an engine in 1824 involving two isotherms and two adiabatic processes, 
Jule studied adiabatic processes, and Clausius formulated the existence of the entropy in the second law of thermodynamics in 1854. 
In 1909 Carath\'eodory made the point regarding the relationship between the connectivity of two states by adiabatic processes and nonintegrability of a distribution, which is defined by the one form of work. Chow proved the general global connectivity in 1934 which was used in studying of partial differential equations. There are significant differences between Riemannian and subRiemannian geometries. However, this geometry can be applied in many situations in our daily life. 
For more details, readers can read the book by Calin and Chang~\cite{CC1}.

A subRiemannian structure over a manifold $\mathcal{M}_n$ is a pair $(\mathcal {D}, \langle\cdot,\cdot\rangle)$, where
$\mathcal{D}$ is a bracket generating distribution and $\langle\cdot,\cdot\rangle$ a fibre inner product defined on
$\mathcal{D}$. The length of the horizontal curve $\gamma$ is
\[
\ell(\gamma):=\int_0^\tau\sqrt{\langle\dot\gamma(s),\dot\gamma(s)\rangle}ds.
\]
The shortest length $d_{cc}(A,B)$ is called a {\it Carnot-Carath\'eodory distance} between $A,B\in \mathcal{M}_n$ which
is given by
\[
d_{cc}(A,B):=\inf\ell(\gamma)
\]
where the infimum is taken over all absolutely continuous horizontal curves joining $A$ and $B$~\cite{CCT}. 

Here we mention two examples of subRiemannian geometry. Let $X_1=\frac{\partial}{\partial x}$ and $X_2=x^m\frac{\partial}{\partial y}$ be the the Grushin vector fields~\cite{CCFI},~\cite{CL1} in $\mathbf R^2$ which satisfy bracket generating property, {\it i.e.,} 
\[
[X_1,[X_1,\ldots, [X_1,X_2]\cdots]]=m!\frac{\partial}{\partial y}.
\]
These vector fields can be used to described parallel parking and even self-driven cars~\cite{CLY}. 
\smallskip 

\subsection{Horizontal Connectivity}
In outcome prediction task of drug prescription, one of the main difficulties is to overcome the distinguishing features under short-term and long-term disease progression.  Moreover, for long term diseases, we need to avoid some low-effective (or even useless or dangerous) drugs. 
The answer to this question not only helps us better characterize the embedding space inferred by Euclidean and subRiemannian distances, but also leads to different optimization formulations.  
Mathematically, the first task is to address the following question: {\it Given any two points on a topologically connected subRiemannian manifold, under what conditions can we join them by a horizontal curve?} In the outcome prediction task of drug prescription, we must distinguish features under short-term and long-term disease progression.  The answer to this question not only helps us to better characterize the embedding space inferred by Euclidean and cosine distance, but also lead to useful optimization formulation under their embedding properties.

To answer this question, we need to prove the following two results. Readers can find more detailed discussions in the book by Calin and Chang~\cite{CC1}. 
\smallskip
 
\noindent {\bf Proposition 3.1.}
Let $U\subset {\mathbf R}^n$ be an open set and $\mathcal D\,=\, {\mbox{span}}\{X_1,\ldots, X_m\}$ be a differentiable distribution on $U$. Then for any point $p\in U$ there is a manifold $V_p^m$ such that 
\smallskip

\noindent $(1)$. $p\in V_p^m$;
\smallskip

\noindent $(2)$. $\dim p\in V_p^m\,=\, m$;
\smallskip

\noindent $(3)$. any two points of $V_p^m$ can be joined by a piecewise horizontal curve.
%\end{prop}
\smallskip

\begin{proof} Let $X_j=\sum_{k=1}^n a_{jk}\partial_{x_k}$ be the vector fields in local coordinates. Consider the ODE system 
\begin{equation}
\label{eq:connect2-1-1} 
\frac{dx_\ell (t)}{dt}\,=-\, F_\ell \big(x(t),u\big),\qquad \ell\,=\, 1,\ldots, n,
\end{equation} 
where $F_\ell\big(x(t),u\big)\,=\, \sum_{j=1}^m a_{j\ell}\big(x(t)\big) u_j$ with $(u_1,\ldots, u_m)\in {\mathbf R}^m$, is a system with $m$ parameters. 

The solutions of~\eqref{eq:connect2-1-1} are horizontal curves with controls $u_j$. Let $x_\ell(0)=x_{\ell 0}$ be the initial conditions of system~\eqref{eq:connect2-1-1}. 
Standard theorems of ODE system provide the existence and local uniqueness of the solutions, which can be expressed by 
\[
x_\ell(t)\,=\, \phi(t; \mathbf x_0;u)\,=\, \phi_\ell\big(t; x_{10},\ldots, x_{n0}; u_1,\ldots, u_m\big),
\]
for $|t|<\varepsilon$, with $\phi_\ell(0;\mathbf x_0;u)=\mathbf x_0$. Since the vector components $a_{j\ell}$ are differentiable, a general theorem states that the functions $\phi_\ell$ 
are twice differentiable with respect to $t$ and locally continuous differentiable with respect to $\mathbf x_0$. 

Since system~\eqref{eq:connect2-1-1} is autonomous, a simple application of he chain rule shows that the functions $\phi_\ell$ verify the relations 
\[
\phi_\ell\big(tt_0;\mathbf x_0;u\big)\,=\, \phi_\ell \big(t_0;\mathbf x_0;tu\big)\,=\, \psi_\ell \big(\mathbf x_0;tu\big),
\]
where $|t_0|<\varepsilon$ and $|tu|<\varepsilon$. 

Applying the theorem on differentiability with respect to a parameter to system~\eqref{eq:connect2-1-1} yields that $\phi_\ell$ are continuous differentiable 
with respect to $u_\ell$ if $|u_\ell|<\delta$ with sufficiently small $\delta>0$. 

If we let $tu_\ell\,=\,\lambda_\ell$, for $\ell\,=\, 1,\ldots, m$, then the formulas 
\[
x_\ell\,=\, \psi_\ell \big(\mathbf x_0;\lambda_1,\ldots, \lambda_m\big),\qquad \ell\,=\, 1,\ldots, n,
\]
for $|\lambda_j|<\varepsilon\delta$ define a $m$-dimensional manifold $V_p^m$ passing through the point $(x_{10},\ldots, x_{n0})$. 
To finish the proof we will need to show that the rank of Jacobian $\frac {\partial \psi_\ell}{\partial \lambda_j}$ is maximum, {\it i.e.,} equal to $m$.
This is equivalent with the fact that the vector fields 
\[
\frac{\partial\psi_1}{\partial\lambda_1},\,\frac{\partial\psi_1}{\partial\lambda_2},\,
\cdots,\, \frac{\partial\psi_n}{\partial\lambda_m}
\] 
are linearly independent. Since $\lambda_\ell=tu_\ell$, it suffices to show that 
\[
\frac{\partial\phi_1}{\partial u_1},\,\frac{\partial\phi_1}{\partial u_2},\,\cdots,\, \frac{\partial\phi_n}{\partial u_m}
\] 
are linearly independent. Since 
\[
\phi\big(t;\mathbf x_0;u\big)\,=\, \exp\Big(t\sum_{k=1}^m u_kX_k\Big),
\]
it follows that 
\[
\frac{\partial\phi}{\partial u_\ell}\,=\, tX_\ell,\qquad \ell\,=\, 1,\ldots, m,
\]
which are linearly independent vector fields for $t\not=0$. It follows that 
\[
{\mbox{rank}}\Big(\frac{\partial\psi_\ell}{\partial\lambda_k}\Big)\,= \, m.
\] 
The proof of this proposition is therefore complete.
\end{proof}
\smallskip

\noindent {\bf Proposition 3.2.}
Let $\mathcal D$ be a nonintegrable distribution. Assume that through each point of the domains $\mathcal R\subset U$ passes a 
$\mathcal D$-connected $m$-dimensional 
manifold $V^m$ defined by the equations
\begin{equation}
\label{eq:connect2-1-8} 
x_j\,=\, f_j\big(\xi_1,\ldots, \xi_m;c_1,\ldots, c_{n-m}\big),\qquad j\,=\, 1,\ldots, n,
\end{equation} 
where $f_j$ are continuous differentiable functions on a domain $\mathcal R^\ast \subset {\mathbf R}^n$, such that 
\[
\frac{\partial(f_1,\ldots, f_n)}{\partial (\xi_1,\ldots, \xi_m;c_1,\ldots, c_{n-m})}\,\not=\, 0\qquad {\mbox{on}}\quad \mathcal R^\ast. 
\]
Then there is a domain $\mathcal R_1\subset \mathcal R^\ast$ such that 
\smallskip

\noindent $(1)$. for all $\mathcal R_1$, there is a $\mathcal D$-connected $(m+1)$-dimensional manifold $V_p^{m+1}$ passing through $p$;
\smallskip

\noindent $(2)$. the functions that define the manifolds $V_p^{m+1}$ on $\mathcal R_1$ have the same properties as the functions $f_j$'s in~\eqref{eq:connect2-1-8}.
%\end{prop}
\smallskip

\begin{proof} Let $\mathcal D={\mbox{span}}\big\{ X_1,\ldots,X_m\big\}$ be the horizontal distribution and 
\[
\mathcal I=\{\theta_1,\ldots,\theta_{n-m}\}
\]
 be the extrinsic ideal associated with $\mathcal D$. Since the distribution $\mathcal D$ is not integrable, the Pfaff system $\mathcal I$ is not integrable; {\it i.e.,} it cannot have integral manifolds of dimension $m$. 
\smallskip

\noindent {\it Proof of statement $(1)$.} For any $p\in \mathcal R$, there is a horizontal vector $X_p\in \mathcal D_p$ such that $X_p\not\in T_pV_p^m$, {\it i.e.,} not tangent to the manifold $V_p^m$. 

The proof of $(1)$ is by contradiction. Let $p\in \mathcal R$ be a fixed point. Assume that any horizontal vector field $X$ about $p$ is tangent to the manifold $V_p^m$. Then 
\[
X_q\,\in\, T_qV_p^m,\qquad \forall\,\,\, q\,\in\, V_p^m.
\]
Therefore $\mathcal D_q\,\subset\, T_qV_p^m$, and since $\dim(\mathcal D_q)\,=\, \dim( T_qV_p^m)\,=\, m$, it follows that the inclusion is in fact identity; {\it i.e.,} 
$\mathcal D_q\,=\, T_qV_p^m$ for all $q\in V_p^m$. 
Since the one-forms $\theta_j$ vanish on $\mathcal D_q$, it follows that $T_qV_p^m$ is an integral $m$-plane for the Pfaff system $\mathcal I$ and hence $V_p^m$ is an integral manifold for $V_p^m$ is an integral manifold $\mathcal I$, which is a contradiction, because $\mathcal I$ is not integrable. 
Hence we prove the assertion $(1)$, 

Let $p_0\in\mathcal R$ be a point with coordinates $(x_{10},x_{20},\ldots, x_+{n0})$ and $u=X_{p_0}$ be the vector given by $(1)$; {\it i.e.,} $u\in \mathcal D_{p_0}$ and $u\not\in T_{p_0}V_{p_0}^m$. Let $(u_1,\ldots, u_m)\in {\mathbf R}^m$ be such that 
\[
u\,=\, \sum_{k=1}^m u_k X_k(p_0).
\] 
The numbers $u_k$ will be kept constant for the rest of the proof. 
\smallskip

\noindent {\it Proof of statement $(2)$.} The matrix 
\[
M\,=\, \left[\begin{matrix} \frac{\partial f_1}{\partial \lambda_1}&\cdots &\frac{\partial f_n}{\partial \lambda_1}\\
\vdots&\cdots &\vdots\\
\frac{\partial f_1}{\partial \lambda_m}&\cdots &\frac{\partial f_n}{\partial \lambda_m}\\
u_1X_i(p_0)&\cdots &u_mX_m(p_0)
\end{matrix}\right]
\]
has rank $m+1$ at the point $p_0$. 

The first $m$ rows of the matrix $M$ are the components of the coordinate vector fields $\frac{\partial f}{\partial \lambda_j}$ on the manifold $V_{p_0}^m$, which are tangent to $V_{p_0}^m$, linearly independent, and span the tangent space $T_{p_0}V_{p_0}^m$. 
The last row of $M$ has the component of the vector $u$, which is transversal to $T_{p_0}V_{p_0}^m$, so all $m+1$ vectors are linearly independent at 
$p_0$ and hence ${\mbox{rank}}(M)=m+1$.

Since all the elements of the matrix $M$ are continuous functions of the coordinates of the point $p_0$, while $u_j$ are still kept constant, there is a subdomain $\mathcal R^\prime\subset\mathcal R$ such that $p_0\in \mathcal R^\prime$ and ${\mbox{rank}}(M)=m+1$ on $\mathcal R^\prime$. 

From the non-vanishing Jacobian condition $\frac{\partial f}{\partial(\lambda, c)}\not=0$ on $\mathcal R^\prime$ it follows that the following $n$ vector fields 
\begin{equation}
\label{eq:connect2-1-9} 
\frac{\partial f}{\partial \lambda_1},\,\ldots, \, \frac{\partial f}{\partial \lambda_m}, \, 
\frac{\partial f}{\partial c_1},\,\ldots, \, \frac{\partial f}{\partial c_{n-m}}
\end{equation} 
are linearly independent on $\mathcal R^\prime$. 

From the preceding discussion, the following $m+1$ vector fields 
\begin{equation}
\label{eq:connect2-1-10} 
\frac{\partial f}{\partial \lambda_1},\,\ldots, \, \frac{\partial f}{\partial \lambda_m}, \, X
\end{equation} 
are linearly independent on $\mathcal R^\prime$. We can complete system~\eqref{eq:connect2-1-10}  with $n-(m+1)$ elements of set~\eqref{eq:connect2-1-9}, say 
\begin{equation}
\label{eq:connect2-1-11} 
\frac{\partial f}{\partial \lambda_1},\,\ldots, \, \frac{\partial f}{\partial \lambda_m}, \, X,\, 
\frac{\partial f}{\partial c_1},\,\ldots, \, \frac{\partial f}{\partial c_{n-m-1}}
\end{equation} 
are linearly independent on $\mathcal R^\prime$. 

In the following we shall deal with the construction of a $(m+1)$-dimensional manifold passing through $p_0$, which depends on $n-(m+1)$ parameters. 
In equation~\eqref{eq:connect2-1-8} consider the parameters $c_{n-m}$ frozen. Let $x_{j0}$ be the coordinates on this new manifold $V^m$. Then 
\begin{equation}
\label{eq:connect2-1-12} 
x_{j0}\,=\, g_j\big( \lambda_1, \lambda_2,\ldots,  \lambda_m; c_1,\ldots, c_{n-m-1}\big),
\end{equation} 
where $g_j$ is continuous differentiable with respect to $\lambda_\ell$ and $c_j$ and 
\[
\frac{\partial g_\ell}{\partial \lambda_\sigma}\,=\, \frac{\partial f_\ell}{\partial \lambda_\sigma},\quad 
\frac{\partial g_\ell}{\partial c_k}\,=\, \frac{\partial f_\ell}{\partial c_k},\,\,\, \sigma=1,\ldots, m,\,\,\, k=1,\ldots, n-m-1.
\]
The equation of the integral curves of the vector field $X$ on $\mathcal R^\prime$ are given by 
\begin{equation}
\label{eq:connect2-1-13} 
x_{j}\,=\, \phi_j\big( t; x_{10}, x_{20},\ldots,  x_{n,0}; u_1,\ldots, u_m\big),\quad j=1,\ldots, n.
\end{equation} 

We shall construct a $(m+1)$-dimensional manifold by pushing the manifold $V^m$ in the direction of the integral curves of $X$. This can be done by substituting the variables $x_{j0}$ given by~\eqref{eq:connect2-1-12} into the expressions provided by~\eqref{eq:connect2-1-13}. 
Let $\lambda_{m+1}=t$ be the $m+1$ variable. We obtain 
\begin{equation*}
\begin{split}
x_\ell \,=&\, \phi_\ell\big(t; \mathbf x_0;u_1,u_2,\ldots, u_m\big)\\
=&\, \phi_\ell\big(\lambda_{m+1}; g_\ell (\lambda_1,\ldots, \lambda_m;c_1,\ldots, c_{n-m-1}); u_1,u_2,\ldots, u_m\big)\\
=&\, F_\ell\big(\lambda_1,\ldots, \lambda_{m+1}; c_1,\ldots, c_{n-m-1})\big)\,\quad \ell=1,\ldots, n,
\end{split}
\end{equation*}
where $u_j$ are kept constant. $F_\ell$ are continuous differentiable functions of $\lambda_1,\ldots, \lambda_{m+1},c_1,\ldots, c_{n-m-1}$.

To show that the equations 
\begin{equation}
\label{eq:connect2-1-14} 
x_\ell \,=\, F_\ell\big(\lambda_1,\ldots, \lambda_{m+1}; c_1,\ldots, c_{n-m-1})\big)
\end{equation} 
defines a manifold of dimension $m+1$, we need to show that 
\begin{equation}
\label{eq:connect2-1-15} 
{\mbox{rank}}\Big( \frac{\partial (F_1,\ldots, F_n)} {\partial( \lambda_1,\ldots, \lambda_{m+1})} \Big)\,=\, m+1
\end{equation} 
on some neighborhood of $p_0$ included in $\mathcal R^\prime$. 

Applying the chain rule yields 
\begin{equation*}
\begin{split}
&\frac{\partial F_\ell}{\partial \lambda_\alpha}\,=\,  \frac{\partial \phi_\ell}{\partial x_{j0}}\frac{\partial x_{j0}}{\partial \lambda_\alpha}\,=\, 
\frac{\partial \phi_\ell}{\partial x_{j0}}\frac{\partial f_j}{\partial \lambda_\alpha}\quad \alpha=1,\ldots, m\\
&\frac{\partial F_\ell}{\partial \lambda_{m+1}}\,=\,  \frac{\partial \phi_\ell}{\partial t}\,=\, u_jX_j\\
&\frac{\partial F_\ell}{\partial c_\beta}\,=\,  \frac{\partial \phi_\ell}{\partial x_{j0}}\frac{\partial x_{j0}}{\partial c_\beta}\,=\, 
\frac{\partial \phi_\ell}{\partial x_{j0}}\frac{\partial f_{j0}}{\partial c_\beta}\quad \beta=1,\ldots, n-(m+1).
\end{split}
\end{equation*}
Since $\det\big [\frac{\partial \phi_\ell}{\partial x_{j0}}\big]\,\not=\, 0$ on a neighborhood of $p_0$, using that vector fields~\eqref{eq:connect2-1-10} are  linearly independent yields 
\[
\frac{\partial F_\ell}{\partial \lambda_1},\, \frac{\partial F_\ell}{\partial \lambda_2},\,\ldots,\,  \frac{\partial F_\ell}{\partial \lambda_m},\, 
\frac{\partial F_\ell}{\partial \lambda_{m+1}}
\]
are linearly independent, which means that~\eqref{eq:connect2-1-15} holds.

Using that~\eqref{eq:connect2-1-11} are linearly independent on $\mathcal R^\prime$, it follows that the vector fields 
\[
\frac{\partial F_\ell}{\partial \lambda_\alpha},\,\frac{\partial F_\ell}{\partial \lambda_{m+1}},\,  \frac{\partial F_\ell}{\partial c_\beta},
\quad \alpha=1,\ldots, m,\,\,\, \beta=1,\ldots, n-(m+1),
\]
are linearly independent on a subdomain $\mathcal R_1\subset \mathcal R^\prime$, which contains $p_0$. Therefore
\[
\frac{\partial (F_1,\ldots, F_n)} {\partial( \lambda_1,\ldots, \lambda_{m+1},c_1,\ldots, c_{n-m-1})} \,\not=\, 0\quad {\mbox{on}}\,\,\, \mathcal R_1, 
\]
and hence the functions $F_\ell$ have the same properties as the functions $f_j$ in~\eqref{eq:connect2-1-8}.

In conclusion, through each point of $\mathcal R_1$ passes a $\mathcal D$-connected $m+1$ manifold defined by equations~\eqref{eq:connect2-1-14}, and each manifold depends on $n-(m+1)$ parameters. We finish the proof of this proposition.

\end{proof}

Now we are in a position to prove the local connectivity property. This result was proved by Teleman  \cite{Tele} for the Pfaff systems that do not contain integrable combinations in 1957. 
Here we shall prove it from the point of view of distributions.  
\smallskip

\noindent {\bf Theorem 3.3.}
\label{thm:Tele} Let $\mathcal D$ be a nonintegrable differentiable distribution of rank $m$ on the open set $U\subset {\mathbf R}^n$.
Then any domain $U_1\subset U$ contains a subdomain $U_2\subset U_1$ such that for any $p$, $q\,\in \, U_2$, there is a piecewise horizontal curve that joins the points $p$ and $q$. 
\smallskip

\begin{proof} From Proposition 3.1, for any $p\in U_1$, there is a $m$-dimensional $\mathcal D$-connected manifold $V_p^m$ passing through $p$. Applying Proposition 3.2 $n-m$ times yields a subdomain $U_2\subset U_1$ such that for all $p\in U_2$, thee is an $n$-dimensional 
$\mathcal D$-connected manifold $W_p^n$ passing through $p$.

Let $p,q\,\in \, U_2$ be two arbitrary points. Let $\gamma$ be a path joining $p$ and $q$ contained in $U_2$ ($\gamma$ not necessarily supposed to be a horizontal curve.) Since $\sup_{x\in \gamma} W_x^n$ covers the compact set ${\mbox{Image}}(\gamma)$, there is a finite subcovering; {\it i.e.,} we can choose $\ell+1$ points on $\gamma$
\[
x_0\,=\, p,\quad x_1,x_2,\ldots, x_{\ell-1},\quad x_\ell=q
\]
such that 
\[
{\mbox{Image}}(\gamma)\,\subset\, \bigcup_{k=0}^\ell W_{x_k}^n.
\]
We can choose the points $x_i$ such that any two consecutive points $x_i$ and $x_{i+1}$ belong to the same manifold $W_{x_i}^n$. Since 
the manifolds $W_{x_i}^n$ are $\mathcal D$-connected , the points $x_i$ and $x_{i+1}$ can be joined by a horizontal curve. This way, the points $p$ and $q$ can be joined by a piecewise horizontal curve. 
\end{proof}
\smallskip

\subsection{Subelliptic Heat Kernel} 
Now we need to use Hamilton or Lagrange formalisms to construct the fundamental solution of the subelliptic heat operator. 
In other words, we are interested in finding the solving kernels for the
operators $\frac{\partial}{\partial t}-\Delta_{\mathcal X}$ Inspired by the Gaussian, it is reasonable
to expect the kernel has the form:
\[
{\mathcal P}_t(x,x_0)=\frac{c}{t^\alpha}e^{-h(x,x_0)}\qquad{\mbox{for some suitable $\alpha$}}. 
\]
See {\it e.g.,} ~\cite{CCFI}, ~\cite{CL1} and ~\cite{CMV}.

The {\it modified complex action} function $h(x,x_0)$ can be written as $\frac{f(x,x_0)}{2t}$ which 
plays the role of $\frac12 \frac{d^2(x,x_0)}{2t}$ and satisfies the Hamilton-Jacobi
equation
\[
\frac{\partial h}{\partial t}+H\Big(x_1,\ldots, x_n,\frac{\partial h}{\partial x_1},\ldots, \frac{\partial h}{\partial x_n}\Big)\,=\, 0. 
\]
In general, when we deal with a subelliptic heat operator, the heat kernel will depends on $n-m$ parameters (or Lagrange multipliers) 
$\tau=(\tau_1,\ldots,\tau_{n-m})$. Furthermore, after calculation, one may see that the action function  $h(x,x_0,\tau,t)$ can be written as 
\[
h(x,x_0,\tau,t)\,=\, \frac {1}{t}g(x,x_0,\tau t, 1).
\]
We look for a heat kernel in the form $\frac {c}{t^\alpha}e^{-h}=\frac {C}{t^\alpha}e^{-\frac{g}{t}}$. The heat kernel should not depend on $\tau$. So we use an age old technique to get rid of $\tau$ by summing over it. Since $\tau$ are continuous parameters. Thus we shall look for a heat kernel in the following form: 
\[
{\mathcal P}_t(x,x_0)\,=\, \frac{1}{(2\pi t)^\alpha}\int_{\mathbf R^{n-m}} e^{-\frac{g(x,x_0,\lambda)}{2t}}V(x,x_0,\lambda)d\lambda.
\]
Here $\tau t=\lambda$ and $V(x,x_0,\lambda)$ is so-called {\it volume element} which is an appropriate measure that makes the integral is convergent.  
Now we may apply properties of the heat kernel and reduce the problem to solving the {\it transport equation} to find $V$. Once we obtain $V$, then the index $\alpha$ can be determined which depends on the {\it Hausdroff dimension} of the subRiemannian manifold $\mathcal M_n$. 
When $m=n$, the $\mathcal M_n$ is an $n$-dimensional Riemannian manifold and $\alpha=\frac n2$ where $n$ is the topological dimension of the manifold.
In the case, the volume element is just the zero section that will recover results in elliptic cases. For more details, readers the books~\cite{CC2}, \cite{CCFI}
and a forthcoming research article. 
\medskip

\section{Unified Framework}
\label{unified-model}
To compare how distance metrics affect prediction performance, we present a unified framework for a graph kernel-based drug prescription prediction system in support of a rigorous empirical evaluation. We consider all disease and distance metric configurations for both  data balance and imbalance ratios. Motivated by MGKF and Cross-Global, a hybrid model is formulated to leverage advantages from these two models. Following the same MGKF three-kernel architecture, namely a Weisfeiler-Lehman subtree kernel $K_{wl}$~\cite{wl-kernel}, Temporal topological kernel  $K_{tp}$~\cite{bcb-2019}, and Vertex histogram kernel $K_{vh}$~\cite{vh-kernel}, we generate a fused kernel embedding with distance metric loss from Cross-Global as regularization .

Specifically, three pairwise kernel matrices via the aforementioned graph kernels are constructed. A single representation for a fused kernel embedding is generated through a deep neural network for successive classification. The distance regularization, achieved by contrastive loss ~\cite{cf-loss}, is integrated to combine the power of deep metric learning, and we force a kernel embedding to preserve an optimal distance property. Semantically similar embeddings are encouraged to be closer to each other, and dissimilar further apart in the kernel space. With this setting, kernel embedding is optimized jointly with classification loss and contrastive loss, deriving a single representation with multi-views and selected distance property. We discuss how embedding and prediction performance differs under different distance metrics in the next section.

%\medskip

Given a set $T$ of $N$ patients with their patient graphs $P_{i} \in T$ where $i \in \{1,...,N\}$ and associated class labels $y \in R^{N \times 1}$ where $y_{i} \in \{1,0\}$, we compute their pairwise kernel gram matrices $\mathrm{\textit{{\textbf{K}}}}_{wl} \in R^{N \times N}$, $\mathrm{\textit{{\textbf{K}}}}_{tp} \in R^{N \times N}$, and  $\mathrm{\textit{{\textbf{K}}}}_{vh} \in R^{N \times N}$ by $K_{wl}$, $K_{tp}$, and $K_{vh}$ respectively. Let ${F_{emb}}$ be a deep neural network parameterized by weight $\theta$ and ${F_{sigmoid}}$ as a single layer sigmoid function parameterized by $w$, we defined an unified framework as the following optimization problem:

\begin{equation}
\label{opt_min}
\begin{aligned}
\min_{\theta, w} \quad \underset{\theta}{\mathcal{L}_{contrastive}} + \underset{w}{\mathcal{L}_{crossentropy}}
\end{aligned}
\end{equation}

\begin{equation}
\label{contrastive}
\begin{aligned}
    \mathcal{L}_{contrastive} = \dfrac{1}{|N|}\sum_{i=1}^{N}\sum_{j=1}^{N}(1 - {\mathrm{\textit{\textbf{Y}}}}_{ij}){max(0, \lambda - {\mathrm{\textit{{\textbf{D}}}}}_{ij})}^{2} + {\mathrm{\textit{\textbf{Y}}}}_{ij}{\mathrm{\textit{{\textbf{D}}}}}_{ij}
\end{aligned}
\end{equation}

\begin{equation}
\label{crossentropy}
\begin{aligned}
    \mathcal{L}_{crossentropy} = \dfrac{1}{|N|}\sum_{i=1}^{N}y_{i}log({F_{sigmoid}}_{w}({F_{emb}}_{\theta}({\mathrm{\textit{{\textbf{K}}}}_{wl}}_{i}, {\mathrm{\textit{{\textbf{K}}}}_{tp}}_{i}, {\mathrm{\textit{{\textbf{K}}}}_{vh}}_{i})))
\end{aligned}
\end{equation}

where $\lambda > 0$ is a constant margin threshold, $\mathrm{\textit{\textbf{Y}}}_{ij} = 1$ if $y_{i} = y_{j}$ else $\mathrm{\textit{\textbf{Y}}}_{ij} = 0$, and ${\mathrm{\textit{{\textbf{D}}}}}_{ij}$ is a pairwise distance metric between kernel embedding ${F_{emb}}_{\theta}({\mathrm{\textit{{\textbf{K}}}}_{wl}}_{i}, {\mathrm{\textit{{\textbf{K}}}}_{tp}}_{i}, {\mathrm{\textit{{\textbf{K}}}}_{vh}}_{i})$ calculated by ~\ref{cosine} or ~\ref{euclidean}. 

Let ${F_{emb}}_{\theta}({\mathrm{\textit{{\textbf{K}}}}_{wl}}_{i}, {\mathrm{\textit{{\textbf{K}}}}_{tp}}_{i}, {\mathrm{\textit{{\textbf{K}}}}_{vh}}_{i}) = {k_{emb}}_{i}$ and ${F_{emb}}_{\theta}({\mathrm{\textit{{\textbf{K}}}}_{wl}}_{j}, {\mathrm{\textit{{\textbf{K}}}}_{tp}}_{j}, {\mathrm{\textit{{\textbf{K}}}}_{vh}}_{j}) = {k_{emb}}_{j}$, we also have:

\begin{equation}
\label{cosine}
\begin{aligned}
{d}_{cosine}({k_{emb}}_{i}, {k_{emb}}_{j}) &= 1 - \dfrac{\langle{k_{emb}}_{i}, {k_{emb}}_{j}\rangle}{\lVert {k_{emb}}_{i} \rVert \cdot \lVert {k_{emb}}_{j} \rVert}
\end{aligned}
\end{equation}

\begin{equation}
\label{euclidean}
\begin{aligned}
{d}_{Euclidean}({k_{emb}}_{i}, {k_{emb}}_{j}) &= {\lVert {k_{emb}}_{i} - {k_{emb}}_{j} \rVert}_{2} 
\end{aligned}
\end{equation}
where $d_{cosine}$ is a cosine distance and $d_{Euclidean}$ is the Euclidean distance. As usual, $\langle \cdot \rangle$ is a standard inner product. $d$ in ~\ref{contrastive} can be either $d_{cosine}$ or $d_{euclidean}$. 

The optimization problem in ~\ref{opt_min} can be solved by mini-batch Stochastic Gradient Descent (SGD). Once we find the optimal $\theta$ and $w$, we can perform the prediction. Considering a new incoming patient with patient graph $P_{\hat{i}}$, we compute pairwise kernel matrices ${\mathrm{\textit{{\textbf{K}}}}_{wl}}_{\hat{i}} \in R^{1 \times N}$, $  {\mathrm{\textit{{\textbf{K}}}}_{tp}}_{\hat{i}} \in R^{1 \times N}$, and ${\mathrm{\textit{{\textbf{K}}}}_{vh}}_{\hat{i}} \in R^{1 \times N}$ between $P_{\hat{i}}$ and all patient graphs in $T$. Then, we have the following decision function:

\begin{equation}
{F_{sigmoid}}_{w}({F_{emb}}_{\theta}({\mathrm{\textit{{\textbf{K}}}}_{wl}}_{\hat{i}}, {\mathrm{\textit{{\textbf{K}}}}_{tp}}_{\hat{i}}, {\mathrm{\textit{{\textbf{K}}}}_{vh}}_{\hat{i}})) = \hat{y}
\end{equation}
where $\hat{y}$ is the predicted class label (e.g., success or failure) of $P_{\hat{i}}$.

\medskip

The problem reduces to find good Riemannian or subRiemannian structures to handle a huge and complicated data set under certain constraints. 
In other words, we need to handle the related optimization problem~\ref{opt_min} by finding horizontal vector fields and then construct solving kernel of the heat operator associate to the subelliptic operator $\Delta_{\mathcal X}$. In this paper, we consider Reproducing Kernel Hilbert Space (RKHS) with certain geometric properties derived from Euclidean or cosine distances.

\medskip

\begin{table}[htbp]
  \centering
  \caption{Dataset / Disease statistics}
    \begin{tabular}{|ccccc|}
    \hline
    Disease & \multicolumn{1}{p{5em}}{Number of cases} & \multicolumn{1}{p{5em}}{Number of failure} & \multicolumn{1}{p{5em}}{Number of success} & \multicolumn{1}{p{5em}|}{Failure-success ratio} \\
    \hline
    Urinary tract infection & 1,501,310 & 703,646 & 797,664 & 47\%:53\% \\
    Acute otitis media & 151,522 & 72,264 & 79,258 & 48\%:52\% \\
    Pneumonia & 95,796 & 37,724 & 58,072 & 39\%:61\% \\
    Acute cystitis & 733,119 & 301,902 & 431,217 & 41\%:59\% \\
    Hypertension & 235,695 & 104,936 & 130,759 & 45\%:55\% \\
    Hyperlipidemia & 123,380 & 26,043 & 97,337 & 21\%:79\% \\
    Diabetes & 131,997 & 34,414 & 97,583 & 26\%:74\% \\
    \hline
    \end{tabular}
  \label{statistics}
\end{table}
%\end{document}

\section{Dataset and Evaluation Protocol}
\label{evaluation}
To investigate how distance metric relates to kernel embedding and prediction, we conduct a rigorous empirical evaluation with our proposed unified framework under different data balance-imbalance ratio on a very large-scale real-world EHRs, a subset of the Taiwanese National Health Insurance Research Database (NHIRD)$^3$\footnote{$^3$https://nhird.nhri.org.tw/en/}.

Our sample of the NHIRD contains a 20-plus year, complete, medical history for over one-million randomly sampled patients. The database is provided by Taiwan's National Health Insurance Administration and the Ministry of Health and Welfare. Data are composed of registration files and original claim data for reimbursement to hospitals that participate in the National Health Insurance (NHI) program. The International Classification of Diseases, 9th Revision, Clinical Modification (ICD9-CM) code indicates the disease diagnosed.  A unique identifier is used per drug and can be further linked to the Anatomical Therapeutic Chemical (ATC) code. For privacy purposes, the NHIRD contains no patient personal information such as name, contact information, and exact birth day (e.g., only with birth year and month). Also all identification numbers for patients and hospitals are de-identified in an attempt to prevent possible information leak. Institutional Review Board (IRB) approvals for our research were granted by all associated institutions. 

We select four short-term diseases and three most prevalent chronic diseases in Taiwan. We follow our previous efforts setting an observation window for each type of disease. Refer Table ~\ref{statistics} for complete disease list, data statistics, and outcome observation setup. To validate the claim, cosine distance is superior under data imbalance in chronic disease, in Cross-Global ~\cite{bcb-2020}, and examine such a case on short-term disease in MGKF ~\cite{bcb-2019}, we prepare balance and imbalance data. For balanced one, we downsample the size of majority cases to minority cases designating rare diseases, while keeping 70 percent of majority cases and 30 percents of minority cases for imbalanced one. The pairwise t-test with a p-value set to 0.01 is used to reject the null-hypothesis to measure the statistical significance for comparisons.

We compare Euclidean (Euclidean) and cosine (Cosine) distance on our unified framework. In ${F_{emb}}$, we set 5000 dimensions for the first embedding layer of each kernel and 50 dimensions for kernel fusion layer, as a two-layer Multi-layer perceptron (MLP). We set 50 dimensions for sigmoid classifier ${F_{sigmoid}}$. During training, we use the Adamax optimizer ~\cite{adam} with a fixed learning rate 0.0001 and setup 64 batch size for 1000 epochs with early stopping criteria on batch loss. Two machine learning models are included as baselines for comparison purposes, e.g., Support Vector Machine (SVM) and Logistic Regression (LR) with all regularization constant set up to 1 (e.g., $C=1$). All patients are represented as documents containing all medical codes from all visits, and transfer to low-dimensional embedding via Paragraph to Vector ~\cite{p2v} with embedding size 512. Accuracy (ACC), Macro F1-score (F1), and the area under the receiver operating characteristic curve (AUC) are used as our evaluation metrics. All models are developed by Tensorflow and scikit-learn packages using Python. The experiments are executed on an Intel Core i9 CPU with 64GB memory and one Nvidia Titan-RTX GPU with 24GB memory$^4$~\footnote{$^4$We do not perform hyper-parameters tuning for all models.}. Accuracy comparisons with other learning models, including a diversity of the latest neural configurations, are found in ~\cite{bcb-2019, bcb-2020}

\section{Main Result}

Tables$^{5,6}$~\footnote{$^5$ Shaded regions indicate statistical equivalence (light) and significant difference (dark) of the Euclidean and cosine measures.}~\footnote{$^6$ A $^\star$ designation indicates statistical significance over all baselines (SVM and LR) with p-value at 0.01.}%~\footnote{$^7$ The significant difference level is determined by setting p-value as 0.01.}
~\ref{table-ut},~\ref{table-aom},~\ref{table-bp}, and ~\ref{table-ac} show evaluation results under short-term diseases with data balance-imbalance, under different models. Chronic diseases are reported in Tables$^{5,6}$ ~\ref{table-hypertension},~\ref{table-hyperlipidemia}, and ~\ref{table-diabetes}. Under a balanced setting, both Euclidean and cosine distance are relatively similar in their short-term and chronic disease evaluations. Euclidean distance even outperforms cosine distance in 5 out of 7 diseases, implying that Euclidean distance can achieve favorable results when data variation is small, no matter for short/long term disease progression. When it comes to an imbalance setting, the Euclidean distance measure is superior to the cosine measure for all short-term diseases.  This confirms our premise that Euclidean distance is applicable to local problems, namely short disease progression. On the other hand, cosine distance is preferable in imbalance long-term chronic disease, which outperforms Euclidean distance especially in F1 score (e.g., an indicator to model performance under imbalance data set). It is worth noting that the evaluation margin between Euclidean and cosine distance is pretty large in all chronic diseases. The degree of outcome variation (e.g., co-morbidity) of long-term chronic disease patient group is larger than patient group in short-term disease, which reflects that Euclidean is more applicable under a low-variation data set. The comparison to the baseline models validates our unified framework. % does not ill-perform and the evaluation is legit and meaningful to draw conclusion. 
Note that, we did not perform any hyper-parameters tuning nor customize to any specific disease group. The purpose of this evaluation was strictly to investigate possible conclusions on model behavior under different distance metrics.  %, and it is not feasible to relate to evaluation in the previous MGKF ~\cite{bcb-2019} and Cross-Global ~\cite{bcb-2020}.

\begin{table}[htbp]
\label{ut}
  \centering
  \caption{Evaluation Results for Urinary tract infection.}
    \begin{tabular}{|cccc|}
    \hline
    \multicolumn{4}{|c|}{Urinary tract infection} \\
    \hline
    \hline
    \multicolumn{4}{|c|}{Balanced} \\
    \hline
    Model & ACC   & F1    & AUC \\
    \hline
    \rowcolor{LightGray}
    \cellcolor{white} Euclidean & $^\star$0.6220 $\pm$ 0.0212 & $^\star$0.6186 $\pm$ 0.0229 & $^\star$0.6220 $\pm$ 0.0212 \\
    \rowcolor{LightGray}
    \cellcolor{white} Cosine & $^\star$\textbf{0.6243 $\pm$ 0.0284} & $^\star$\textbf{0.6216 $\pm$ 0.0283} & $^\star$\textbf{0.6243 $\pm$ 0.0284} \\
    SVM   & 0.5047 $\pm$ 0.0257 & 0.5046 $\pm$ 0.0258 & 0.5047 $\pm$ 0.0257  \\
    LR    & 0.5210 $\pm$ 0.0240 & 0.4874 $\pm$ 0.0300 & 0.4988 $\pm$ 0.0296 \\
    \hline
    \multicolumn{4}{|c|}{Imbalanced} \\
    \hline
    \rowcolor{LightGray}
    \cellcolor{white} Euclidean & $^\star$\textbf{0.6280 $\pm$ 0.0469} & $^\star$0.5465 $\pm$ 0.0290 & $^\star$0.5895 $\pm$ 0.0249 \\
    \rowcolor{LightGray}
    \cellcolor{white} Cosine & $^\star$0.6165 $\pm$ 0.0532 & $^\star$\textbf{0.5632 $\pm$ 0.0356} & $^\star$\textbf{0.6194 $\pm$ 0.0182} \\
    SVM   & 0.5023 $\pm$ 0.0266 & 0.5051 $\pm$ 0.0266 & 0.5053 $\pm$ 0.0265  \\
    LR    & 0.5208 $\pm$ 0.0240 & 0.4872 $\pm$ 0.0260 & 0.4896 $\pm$ 0.0297 \\
    \hline
    \end{tabular}
  \label{table-ut}
\end{table}

\begin{table}[htbp]
  \centering
  \caption{Evaluation Results for Acute otitis media.}
    \begin{tabular}{|cccc|}
    \hline
    \multicolumn{4}{|c|}{Acute otitis media} \\
    \hline
    \hline
    \multicolumn{4}{|c|}{Balanced} \\
    \hline
    Model & ACC   & F1    & AUC \\
    \hline
    \rowcolor{LightGray}
    \cellcolor{white} Euclidean & $^\star$\textbf{0.6245 $\pm$ 0.0200} &\cellcolor{LightGray} $^\star$\textbf{0.6224 $\pm$ 0.0218} & $^\star$\textbf{0.6245 $\pm$ 0.0200} \\
    \rowcolor{LightGray}
    \cellcolor{white} Cosine & $^\star$0.6138 $\pm$ 0.0183 &\cellcolor{LightGray} $^\star$0.6097 $\pm$ 0.0185 & $^\star$0.6137 $\pm$ 0.0185 \\
    SVM & 0.5023 $\pm$ 0.0204 & 0.5021 $\pm$ 0.0203 & 0.5023 $\pm$ 0.0203 \\
    LR & 0.5011 $\pm$ 0.0211 & 0.5010 $\pm$ 0.0211 & 0.5011 $\pm$ 0.0212 \\
    \hline
    \multicolumn{4}{|c|}{Imbalanced} \\
    \hline
    \rowcolor{LightGray}
    \cellcolor{white} Euclidean & $^\star$\textbf{0.6570 $\pm$ 0.0203} &\cellcolor{LightGray} $^\star$0.5453 $\pm$ 0.0342 &\cellcolor{LightGray} $^\star$0.6037 $\pm$ 0.0324 \\
    \rowcolor{LightGray}
    \cellcolor{white} Cosine & $^\star$0.6238 $\pm$ 0.0306 &\cellcolor{LightGray} $^\star$\textbf{0.5554 $\pm$ 0.0258} &\cellcolor{LightGray} $^\star$\textbf{0.6042 $\pm$ 0.0346} \\
    SVM   & 0.5165 $\pm$ 0.0196 & 0.4803 $\pm$ 0.0212 & 0.4899 $\pm$ 0.0246 \\
    LR    & 0.5170 $\pm$ 0.0177 & 0.4804 $\pm$ 0.0201 & 0.4898 $\pm$ 0.0237 \\
    \hline
    \end{tabular}
  \label{table-aom}
\end{table}

\begin{table}[htbp]
\label{table-bp}
  \centering
  \caption{Evaluation Results for Pneumonia.}
    \begin{tabular}{|cccc|}
    \hline
    \multicolumn{4}{|c|}{Pneumonia} \\
    \hline
    \hline
    \multicolumn{4}{|c|}{Balanced} \\
    \hline
    Model & ACC   & F1    & AUC \\
    \hline
    \rowcolor{LightGray}
    \cellcolor{white} Euclidean & $^\star$0.6013 $\pm$ 0.0279 & $^\star$\textbf{0.5922 $\pm$ 0.3112} & $^\star$0.6013 $\pm$ 0.0279 \\
    \rowcolor{LightGray}
    \cellcolor{white} Cosine & $^\star$\textbf{0.6023 $\pm$ 0.0211} & $^\star$0.5918 $\pm$ 0.0263 & $^\star$\textbf{0.6023 $\pm$ 0.0211} \\
    SVM   & 0.4976 $\pm$ 0.0127 & 0.4975 $\pm$ 0.0127 & 0.4976 $\pm$ 0.0126 \\
    LR    & 0.4979 $\pm$ 0.0130 & 0.4978 $\pm$ 0.0130 & 0.4979 $\pm$ 0.0129 \\
    \hline
    \multicolumn{4}{|c|}{Imbalanced} \\
    \hline
    \rowcolor{LightGray}
    \cellcolor{white} Euclidean & $^\star$\textbf{0.6398 $\pm$ 0.0688} &\cellcolor{LightGray} $^\star$0.5626 $\pm$ 0.0423 &\cellcolor{LightGray} $^\star$0.6028 $\pm$ 0.0270 \\
    \rowcolor{LightGray}
    \cellcolor{white} Cosine & $^\star$0.6255 $\pm$ 0.0470 &\cellcolor{LightGray} $^\star$\textbf{0.5712 $\pm$ 0.0209} &\cellcolor{LightGray} $^\star$\textbf{0.6220 $\pm$ 0.0250} \\
     SVM   & 0.5430 $\pm$ 0.0243 & 0.5070 $\pm$ 0.0246 & 0.5179 $\pm$ 0.0268 \\
    LR    & 0.5430 $\pm$ 0.0243 & 0.5074 $\pm$ 0.0242 & 0.5186 $\pm$ 0.0261 \\
    \hline
    \end{tabular}
  \label{table-bp}
\end{table}

\begin{table}[htbp]
  \centering
  \caption{Evaluation Results for Acute cystitis.}
    \begin{tabular}{|cccc|}
    \hline
    \multicolumn{4}{|c|}{Acute cystitis} \\
    \hline
    \hline
    \multicolumn{4}{|c|}{Balanced} \\
    \hline
    Model & ACC   & F1    & AUC \\
    \hline
    \rowcolor{LightGray}
    \cellcolor{white} Euclidean & $^\star$\textbf{0.6143 $\pm$ 0.0189} & $^\star$\textbf{0.6087 $\pm$ 0.0245} & $^\star$\textbf{0.6143 $\pm$ 0.0189} \\
    \rowcolor{LightGray}
    \cellcolor{white} Cosine & $^\star$0.6095 $\pm$ 0.0182 & $^\star$0.6068 $\pm$ 0.0199 & $^\star$0.6095 $\pm$ 0.0182 \\
    SVM & 0.5049 $\pm$ 0.0231 & 0.5048 $\pm$ 0.0231 & 0.5049 $\pm$ 0.0231 \\
    LR & 0.5037 $\pm$ 0.0228 & 0.5037 $\pm$ 0.0228 & 0.5037 $\pm$ 0.0228 \\
    \hline
    \multicolumn{4}{|c|}{Imbalanced} \\
    \hline
    \rowcolor{LightGray}
    \cellcolor{white} Euclidean & $^\star$\textbf{0.6353 $\pm$ 0.0346} &\cellcolor{LightGray} $^\star$0.5607 $\pm$ 0.0231 &\cellcolor{LightGray} $^\star$0.5763 $\pm$ 0.0325 \\
    \rowcolor{LightGray}
    \cellcolor{white} Cosine & $^\star$0.6280 $\pm$ 0.0405 &\cellcolor{LightGray} $^\star$\textbf{0.5632 $\pm$ 0.0201} &\cellcolor{LightGray} $^\star$\textbf{0.5839 $\pm$ 0.0267} \\
    SVM   & 0.5235 $\pm$ 0.0227 & 0.4871 $\pm$ 0.0219 & 0.4965 $\pm$ 0.0233 \\
    LR    & 0.5230 $\pm$ 0.0248 & 0.4860 $\pm$ 0.0232 & 0.4957 $\pm$ 0.0242 \\
    \hline
    \end{tabular}
  \label{table-ac}
\end{table}

\begin{table}[htbp]
  \centering
  \caption{Evaluation Results for Hypertension.}
    \begin{tabular}{|cccc|}
    \hline
    \multicolumn{4}{|c|}{Hypertension} \\
    \hline
    \hline
    \multicolumn{4}{|c|}{Balanced} \\
    \hline
    Model & ACC   & F1    & AUC \\
    \hline
    \rowcolor{LightGray}
    \cellcolor{white} Euclidean & $^\star$\textbf{0.7315 $\pm$ 0.0126} & $^\star$\textbf{0.7305 $\pm$ 0.0131} & $^\star$\textbf{0.7315 $\pm$ 0.0126} \\
    \rowcolor{LightGray}
    \cellcolor{white} Cosine & $^\star$0.7290 $\pm$ 0.0131 & $^\star$0.7282 $\pm$ 0.0139 & $^\star$0.7290 $\pm$ 0.0131 \\
    SVM   & 0.4956 $\pm$ 0.0165 & 0.4954 $\pm$ 0.0165 & 0.4956 $\pm$ 0.0165 \\
    LR    & 0.4961 $\pm$ 0.0168 & 0.4959 $\pm$ 0.0168 & 0.4961 $\pm$ 0.0167 \\
    \hline
    \multicolumn{4}{|c|}{Imbalanced} \\
    \hline
    \rowcolor{DarkGray}
    \cellcolor{white} Euclidean & $^\star$0.7036 $\pm$ 0.005 & 0.4255 $\pm$ 0.0180 & 0.5065 $\pm$ 0.0086 \\
    \rowcolor{DarkGray}
    \cellcolor{white} \cellcolor{white} Cosine & $^\star$\textbf{0.7497 $\pm$ 0.0236} & $^\star$\textbf{0.6525 $\pm$ 0.0500} & $^\star$\textbf{0.6455 $\pm$ 0.0418} \\
    SVM   & 0.5163 $\pm$ 0.0223 & 0.4785 $\pm$ 0.0228 & 0.4871 $\pm$ 0.0253 \\
    LR    & 0.5170 $\pm$ 0.0227 & 0.4791 $\pm$ 0.0229 & 0.4876 $\pm$ 0.0254 \\
    \hline
    \end{tabular}
  \label{table-hypertension}
\end{table}

\begin{table}[htbp]
  \centering
  \caption{Evaluation Results for Hyperlipidemia.}
    \begin{tabular}{|cccc|}
    \hline
    \multicolumn{4}{|c|}{Hyperlipidemia} \\
    \hline
    \hline
    \multicolumn{4}{|c|}{Balanced} \\
    \hline
    Model & ACC   & F1    & AUC \\
    \hline
    \rowcolor{LightGray}
    \cellcolor{white} Euclidean & $^\star$\textbf{0.7478 $\pm$ 0.0193} & $^\star$\textbf{0.7469 $\pm$ 0.0197} & $^\star$\textbf{0.7478 $\pm$ 0.0193} \\
    \rowcolor{LightGray}
    \cellcolor{white} Cosine & $^\star$0.7470 $\pm$ 0.0185 & $^\star$0.7461 $\pm$ 0.0187 & $^\star$0.7470 $\pm$ 0.0185 \\
    SVM   & 0.4891 $\pm$ 0.0172 & 0.4890 $\pm$ 0.0171 & 0.4891 $\pm$ 0.0172 \\
    LR    & 0.4888 $\pm$ 0.0181 & 0.4887 $\pm$ 0.0181 & 0.4888 $\pm$ 0.0181 \\
    \hline
    \multicolumn{4}{|c|}{Imbalanced} \\
    \hline
    \rowcolor{DarkGray}
    \cellcolor{white} Euclidean & $^\star$0.7323 $\pm$ 0.0408 & 0.5312 $\pm$ 0.1482 & $^\star$0.5797 $\pm$ 0.1011 \\
    \rowcolor{DarkGray}
    \cellcolor{white} Cosine & $^\star$\textbf{0.7565 $\pm$ 0.0356} & $^\star$\textbf{0.6402 $\pm$ 0.0955} & $^\star$\textbf{0.6415 $\pm$ 0.0730} \\
    SVM   & 0.5183 $\pm$ 0.0275 & 0.4821 $\pm$ 0.0264 & 0.4914 $\pm$ 0.0281 \\
    LR    & 0.5193 $\pm$ 0.0272 & 0.4830 $\pm$ 0.0263 & 0.4923 $\pm$ 0.0281 \\
    \hline
    \end{tabular}
  \label{table-hyperlipidemia}
\end{table}

\begin{table}[htbp]
  \centering
  \caption{Evaluation Results for Diabetes.}
    \begin{tabular}{|cccc|}
    \hline
    \multicolumn{4}{|c|}{Diabetes} \\
    \hline
    \hline
    \multicolumn{4}{|c|}{Balanced} \\
    \hline
    Model & ACC   & F1    & AUC \\
    \hline
    \rowcolor{LightGray}
    \cellcolor{white} Euclidean & $^\star$\textbf{0.7085 $\pm$ 0.0151} & $^\star$\textbf{0.7064 $\pm$ 0.0156} & $^\star$\textbf{0.7085 $\pm$ 0.0151} \\
    \rowcolor{LightGray}
    \cellcolor{white} Cosine & $^\star$0.6973 $\pm$ 0.0191 & $^\star$0.6958 $\pm$ 0.0191 & $^\star$0.6973 $\pm$ 0.0191 \\
    SVM   & 0.4894 $\pm$ 0.0151 & 0.4892 $\pm$ 0.0151 & 0.4894 $\pm$ 0.0151 \\
    LR    & 0.4886 $\pm$ 0.0148 & 0.4884 $\pm$ 0.0147 & 0.4886 $\pm$ 0.0148 \\
    \hline
    \multicolumn{4}{|c|}{Imbalanced} \\
    \hline
    \rowcolor{DarkGray}
    \cellcolor{white} Euclidean & $^\star$0.7025 $\pm$ 0.0075 & 0.4315 $\pm$ 0.0591 & 0.5104 $\pm$ 0.0311 \\
    \rowcolor{DarkGray}
    \cellcolor{white} Cosine & $^\star$\textbf{0.7353 $\pm$ 0.0195} & $^\star$\textbf{0.6096 $\pm$ 0.0684} & $^\star$\textbf{0.6114 $\pm$ 0.0495} \\
    SVM   & 0.5168 $\pm$ 0.0164 & 0.4827 $\pm$ 0.0126 & 0.4939 $\pm$ 0.0133 \\
    LR    & 0.5167 $\pm$ 0.0167 & 0.4828 $\pm$ 0.0125 & 0.4941 $\pm$ 0.0130 \\
    \hline
    \end{tabular}%
  \label{table-diabetes}%
\end{table}%

\begin{figure}
\centering
\hfill
\subfigure[Urinary tract infection training set with Euclidean distance. \label{ut-train-e}]{\includegraphics[scale=0.14]{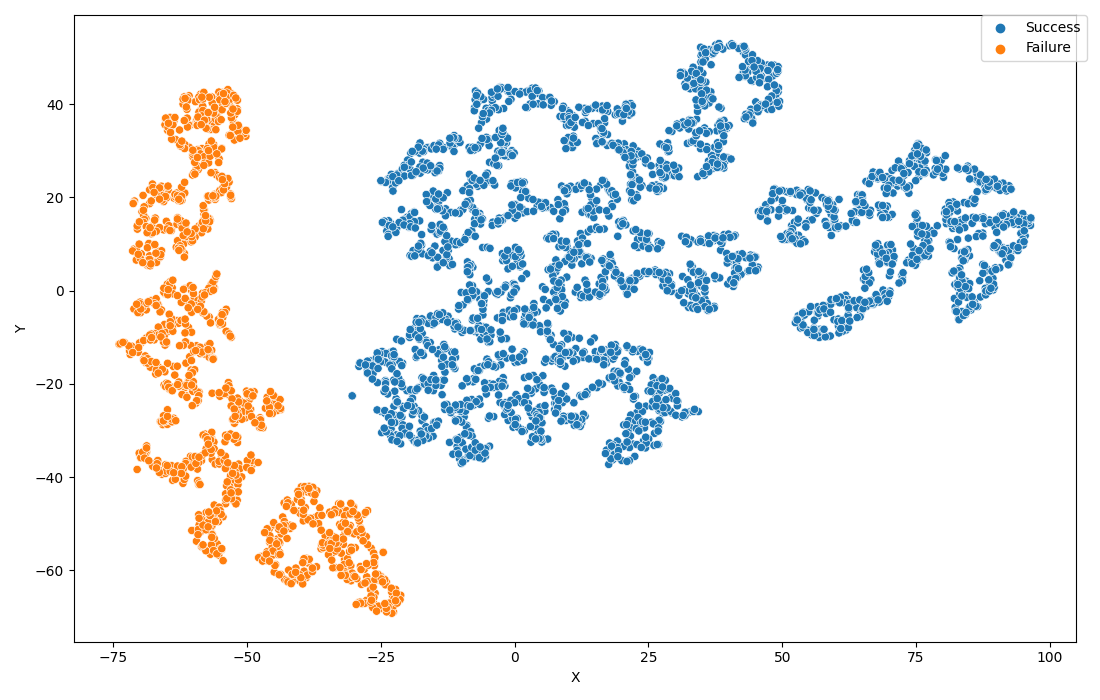}}
\hfill
\subfigure[Urinary tract infection training set with cosine distance. \label{ut-train-c}]{\includegraphics[scale=0.14]{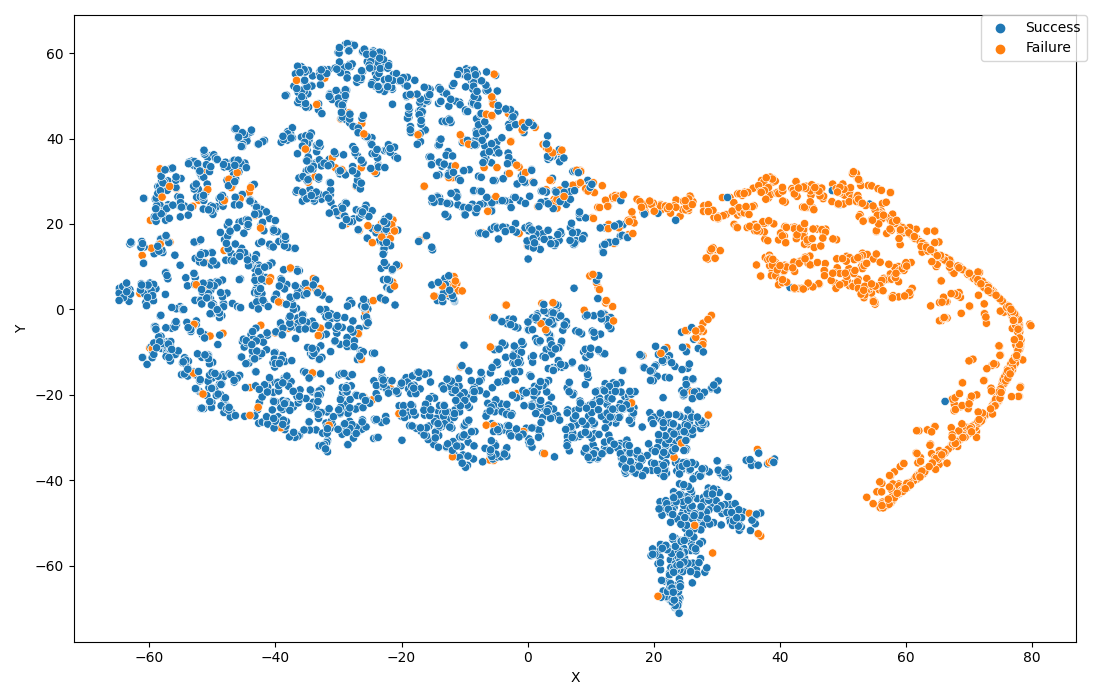}}
\hfill
\subfigure[Urinary tract infection testing set with Euclidean distance. \label{ut-test-e}]{\includegraphics[scale=0.14]{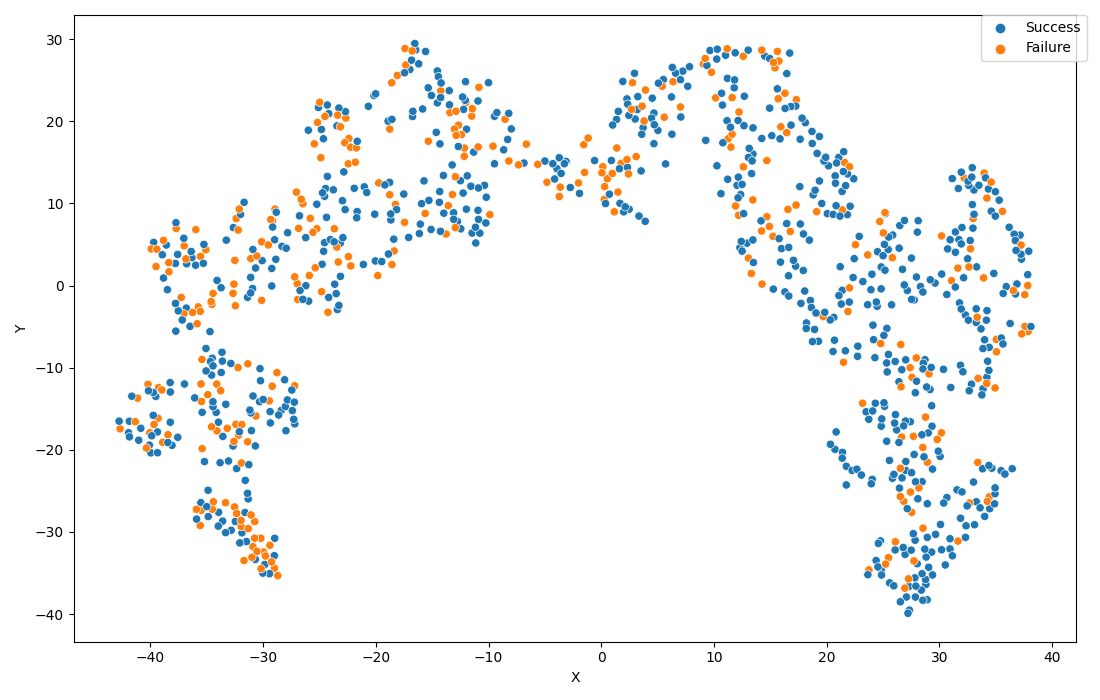}}
\hfill
\subfigure[Urinary tract infection testing set with cosine distance. \label{ut-test-c}]{\includegraphics[scale=0.14]{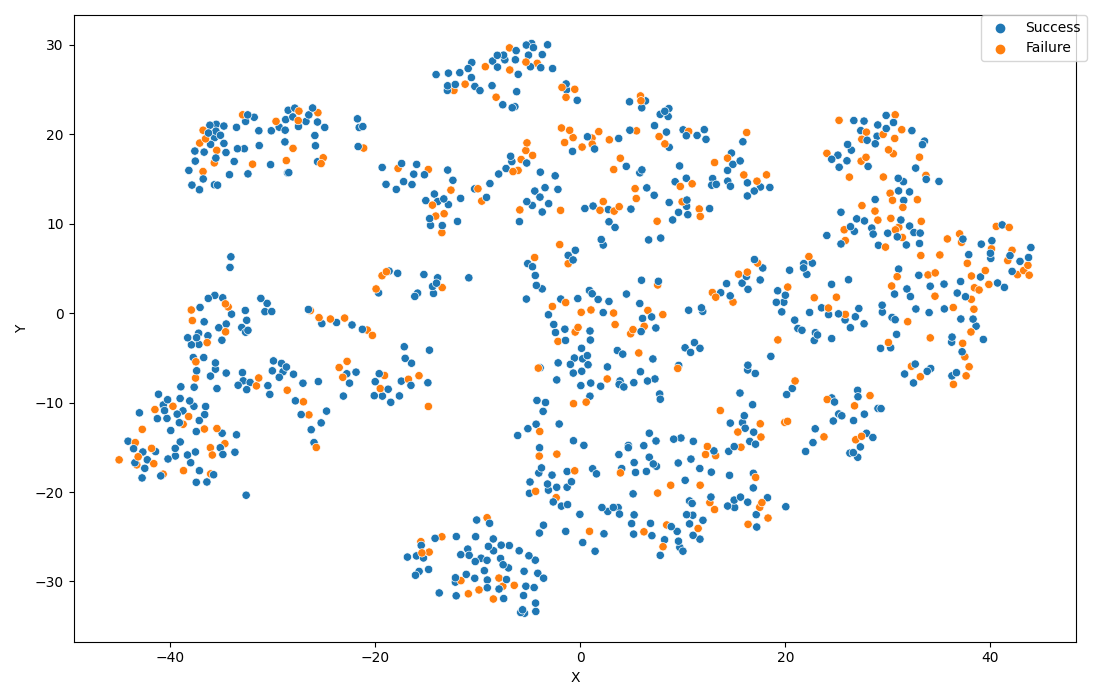}}
\centering
\hfill
\subfigure[Acute otitis media training set with Euclidean distance. \label{aom-train-e}]{\includegraphics[scale=0.14]{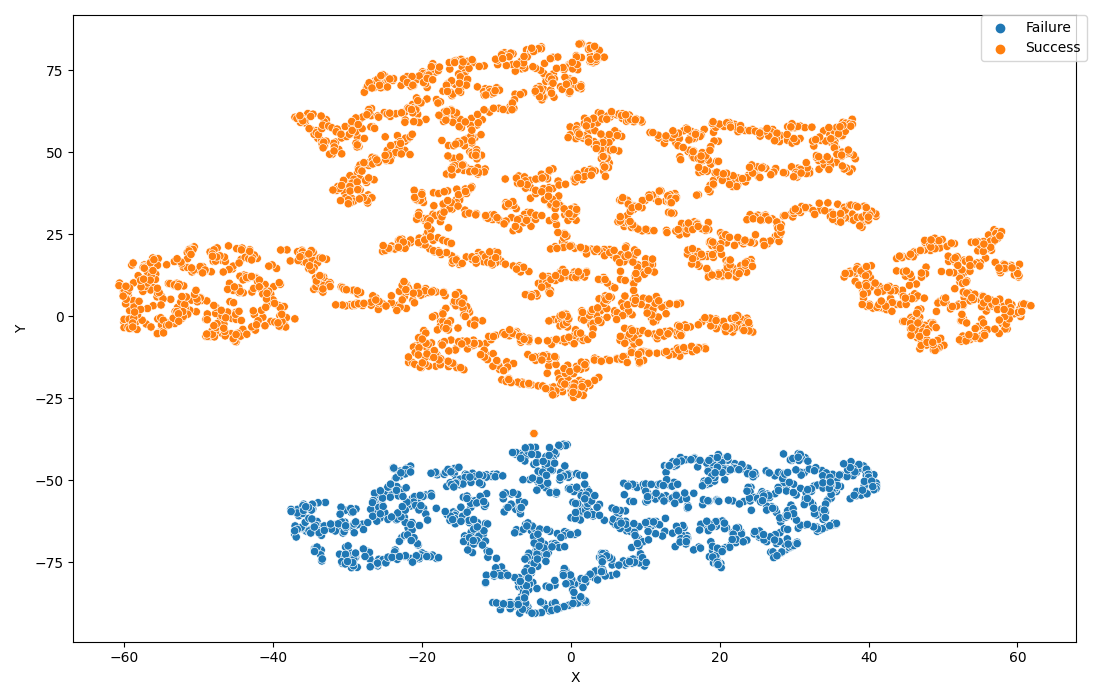}}
\hfill
\subfigure[Acute otitis media training set with cosine distance. \label{aom-train-c}]{\includegraphics[scale=0.14]{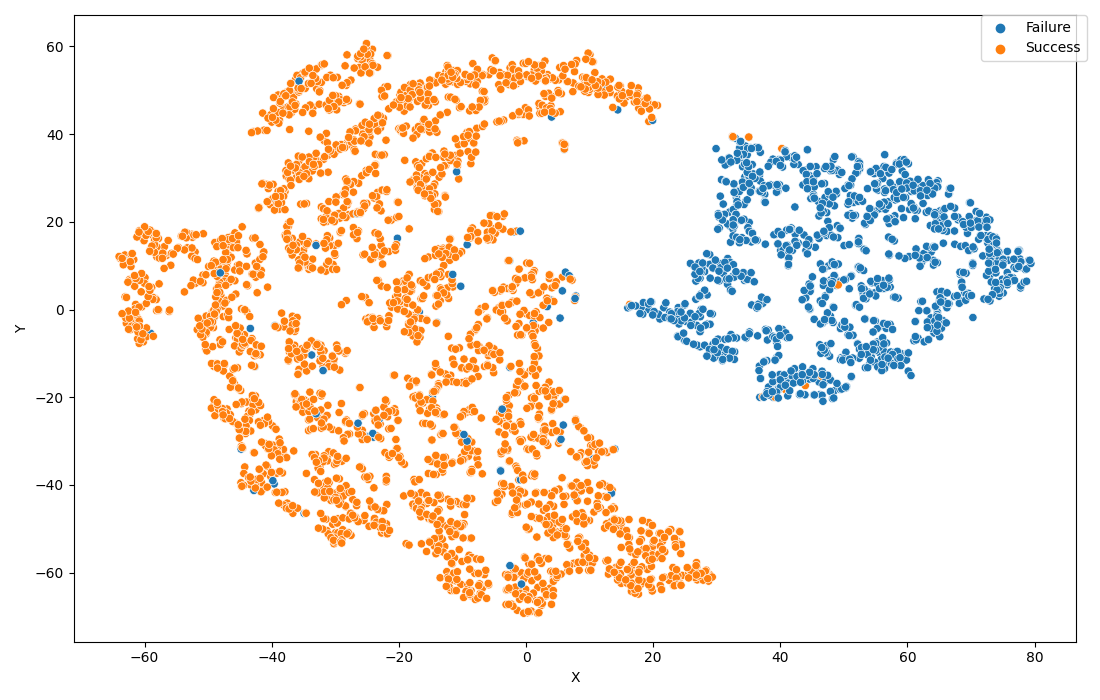}}
\hfill
\subfigure[Acute otitis media testing set with Euclidean distance. \label{aom-test-e}]{\includegraphics[scale=0.14]{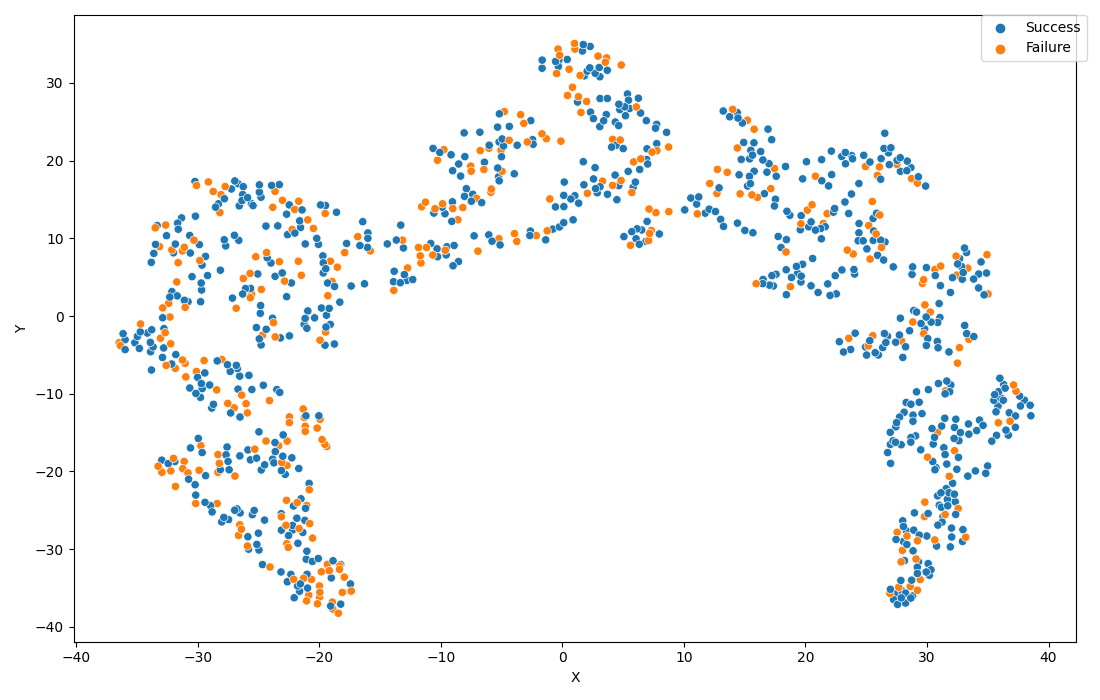}}
\hfill
\subfigure[Acute otitis media testing set with cosine distance. \label{aom-test-c}]{\includegraphics[scale=0.14]{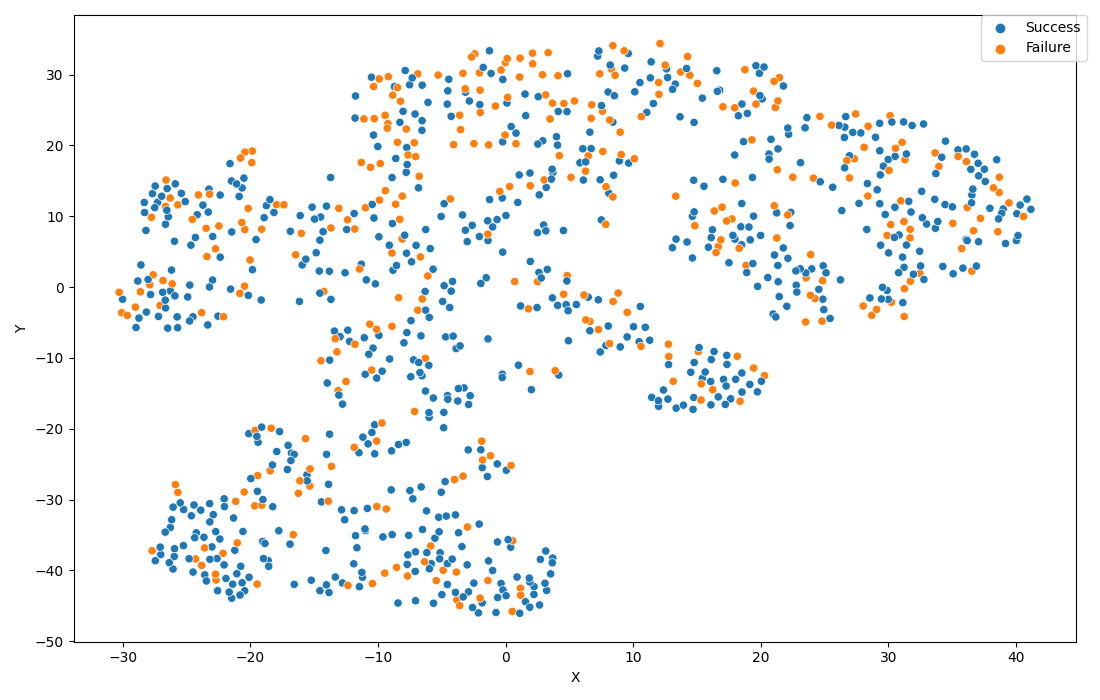}}
\centering
\hfill
\subfigure[Pneumonia training set with Euclidean distance. \label{bp-train-e}]{\includegraphics[scale=0.14]{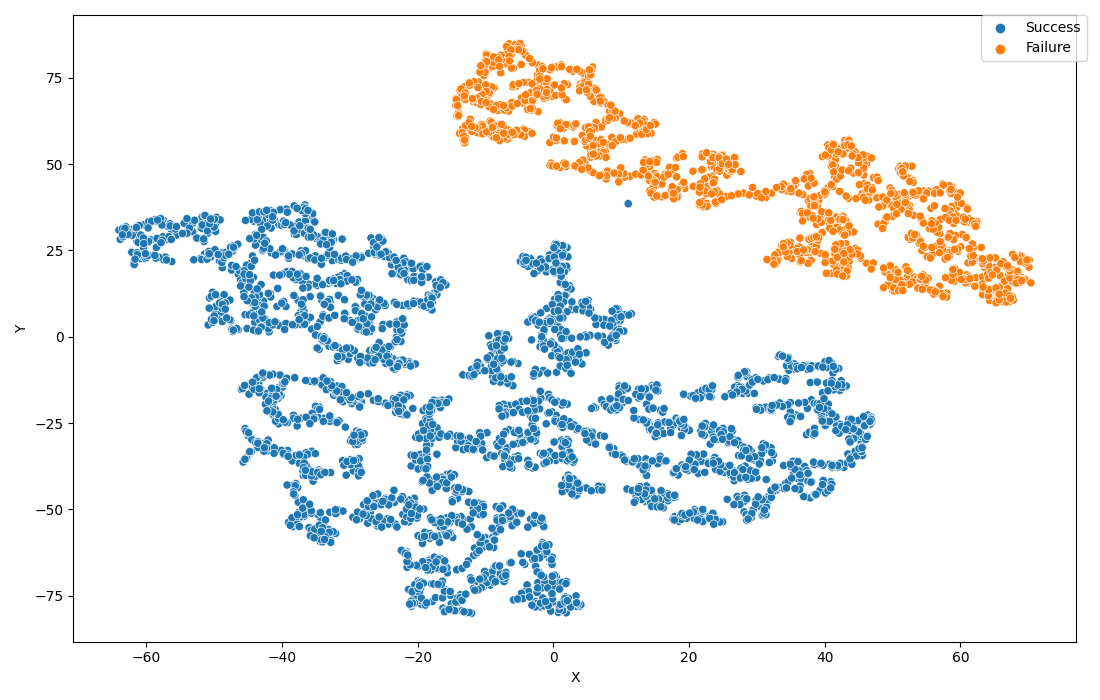}}
\hfill
\subfigure[Pneumonia training set with cosine distance. \label{bp-train-c}]{\includegraphics[scale=0.14]{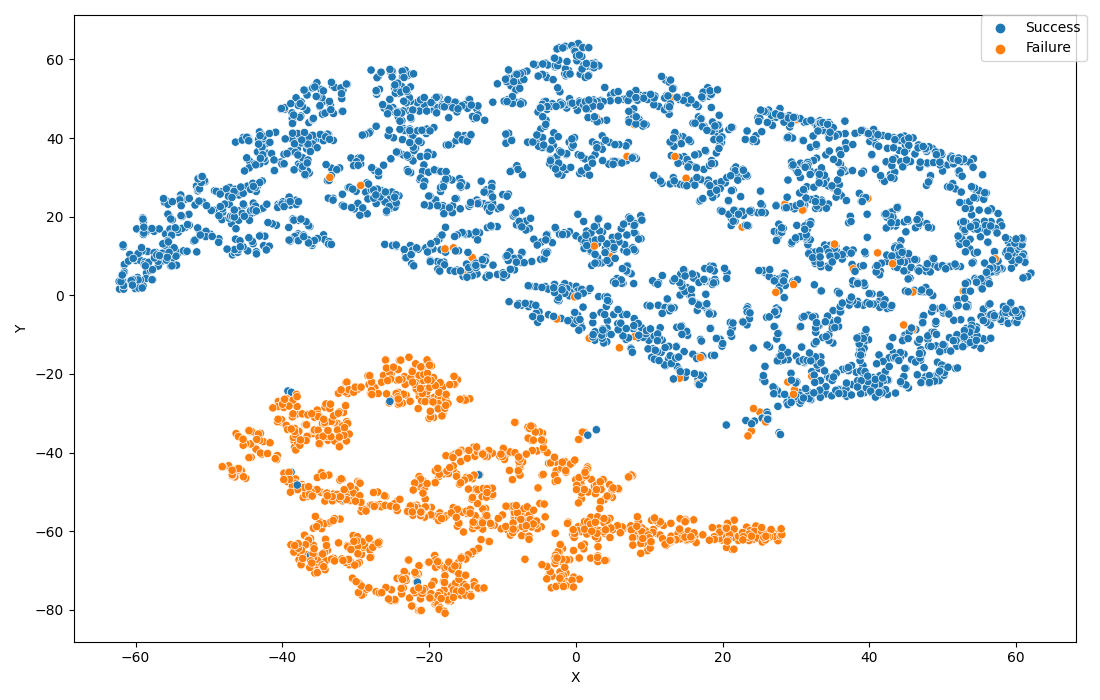}}
\hfill
\subfigure[Pneumonia testing set with Euclidean distance. \label{bp-test-e}]{\includegraphics[scale=0.14]{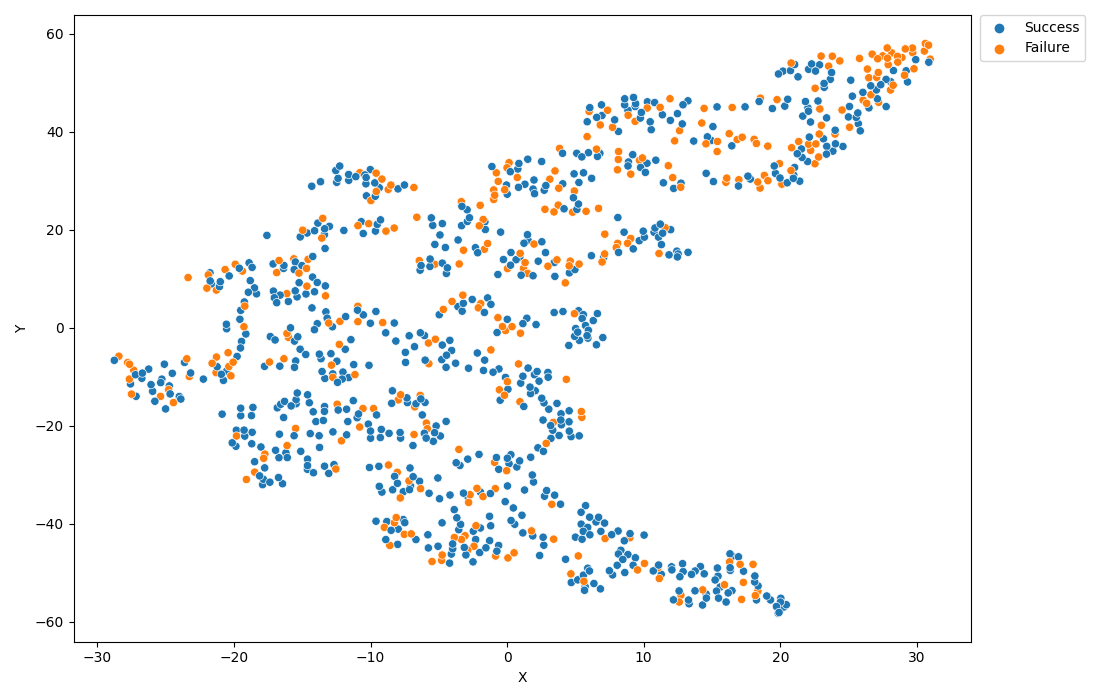}}
\hfill
\subfigure[Pneumonia testing set with cosine distance. 
\label{bp-test-c}]{\includegraphics[scale=0.14]{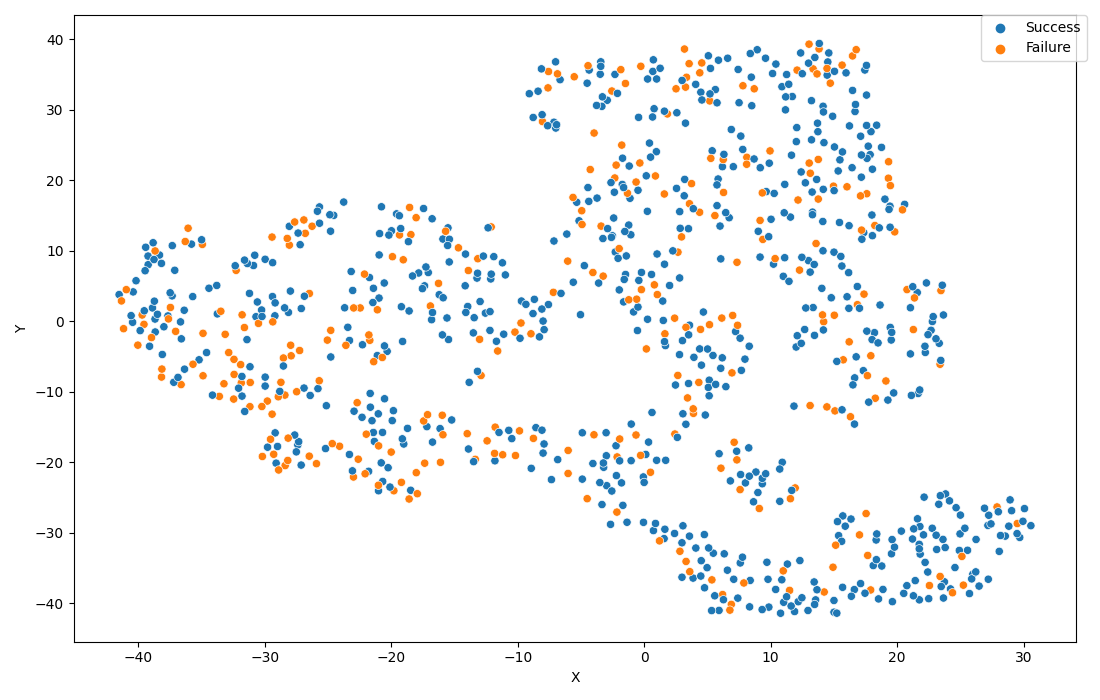}}
\centering
\hfill
\subfigure[Acute cystitis training set with Euclidean distance. \label{ac-train-e}]{\includegraphics[scale=0.14]{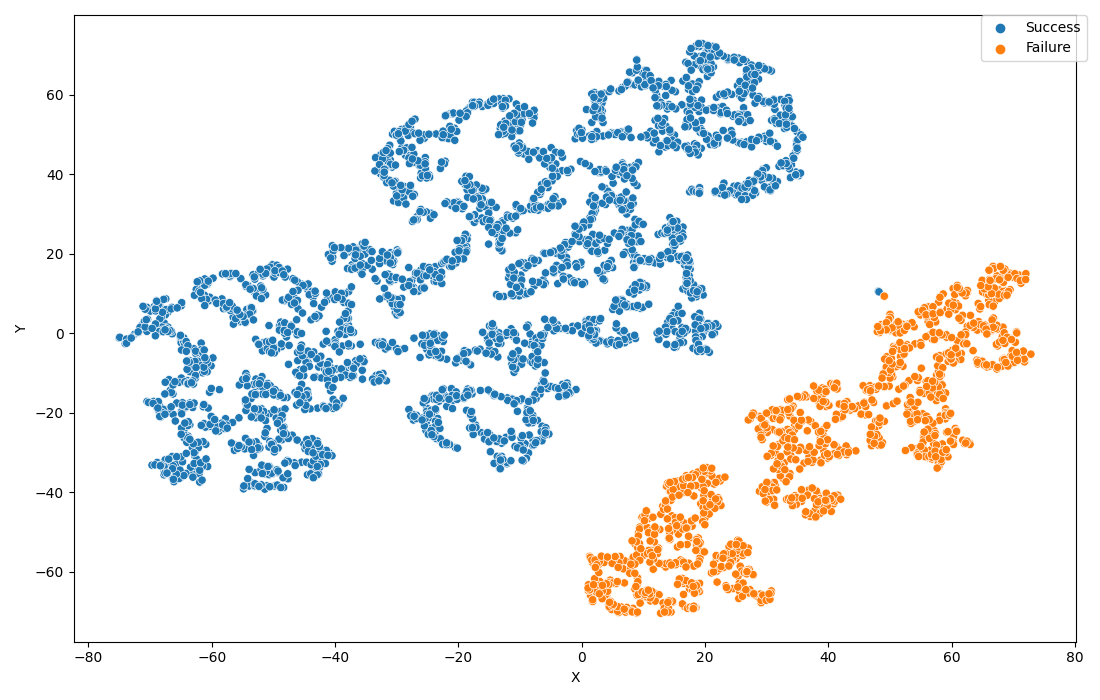}}
\hfill
\subfigure[Acute cystitis training set with cosine distance. \label{ac-train-c}]{\includegraphics[scale=0.14]{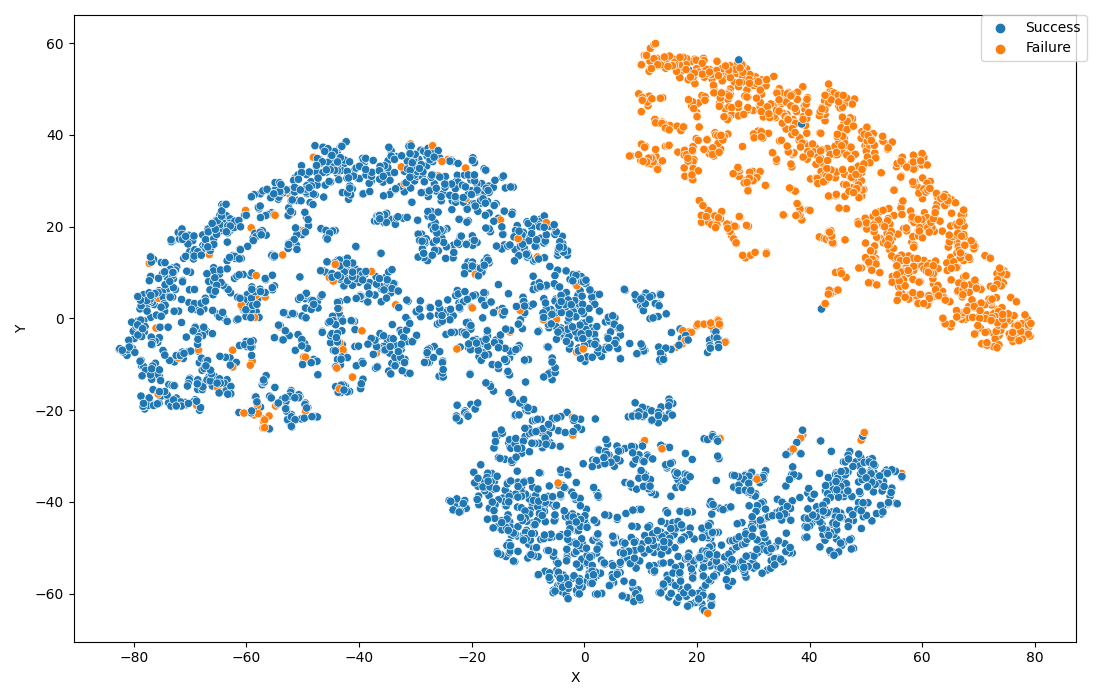}}
\hfill
\subfigure[Acute cystitis testing set with Euclidean distance. \label{ac-test-e}]{\includegraphics[scale=0.14]{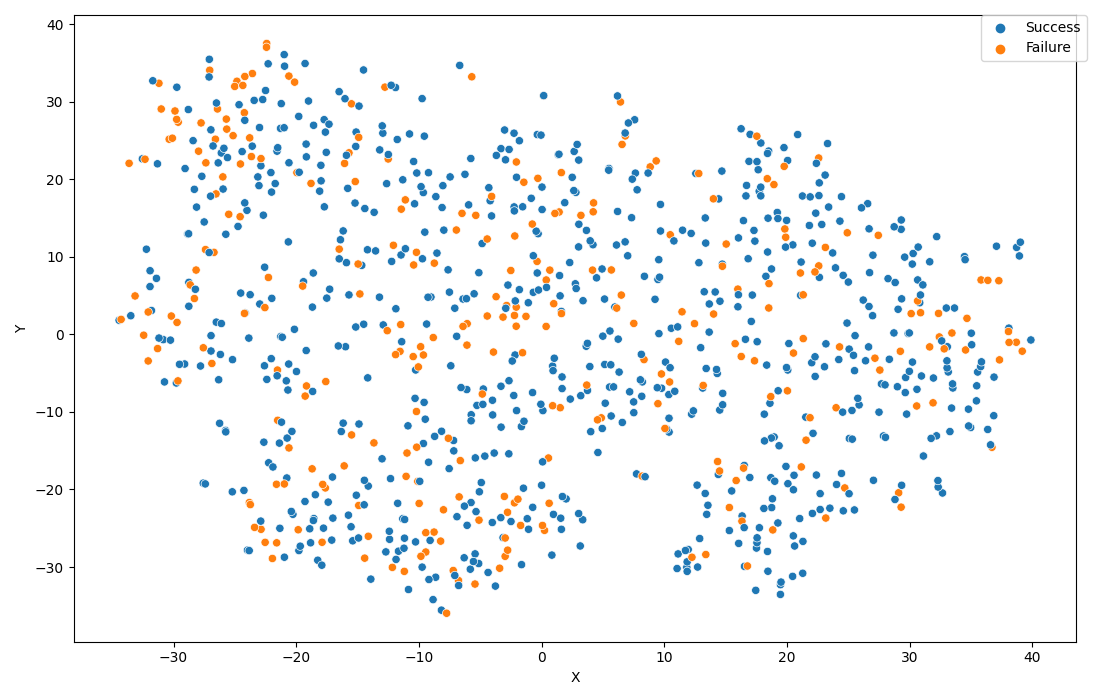}}
\hfill
\subfigure[Acute cystitis testing set with cosine distance. \label{ac-test-c}]{\includegraphics[scale=0.14]{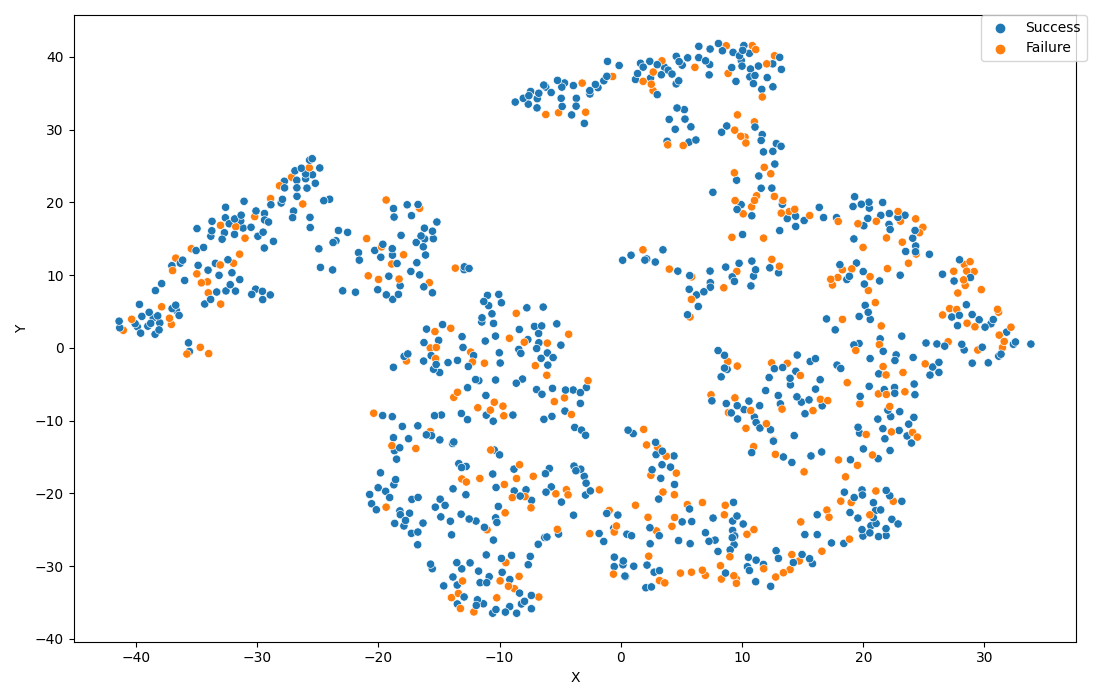}}
\caption{2D visualization of kernel embedding by tSNE with different distance metrics under short-term diseases. We fit our unified framework on training set and evaluate out-of-sample embedding ability on testing set. Blue dot denotes failure case, and orange dot denotes success case.}
\label{short-term-embedding}
\end{figure}

\begin{figure}
\centering
\hfill
\subfigure[Hypertension training set with Euclidean distance. \label{hypertension-train-e}]{\includegraphics[scale=0.14]{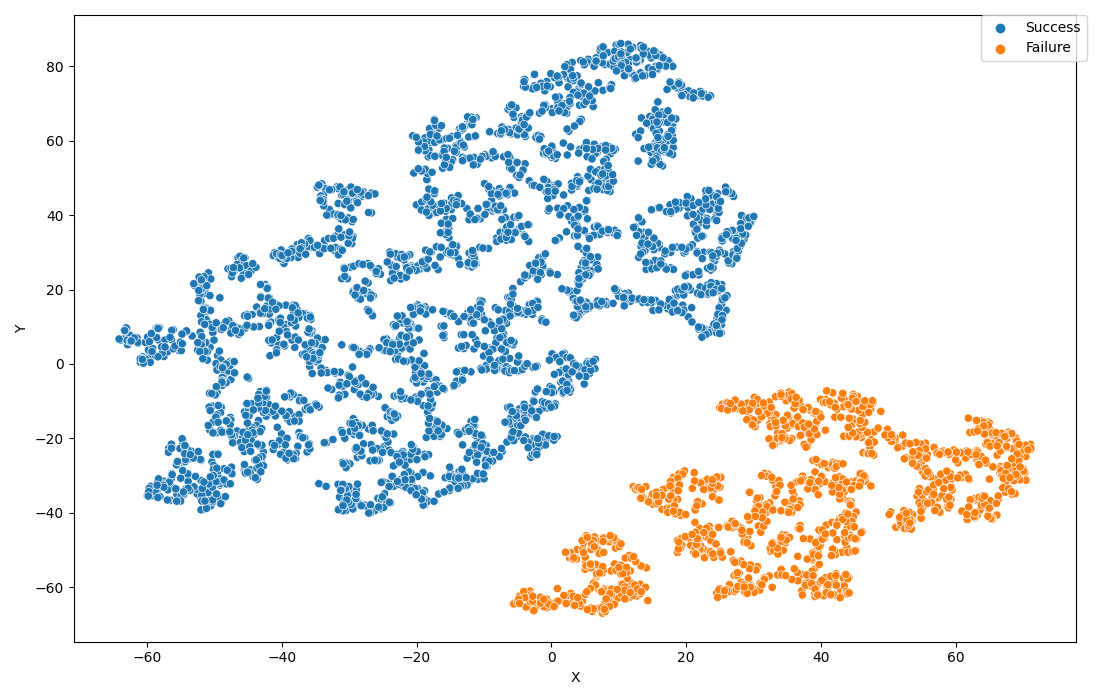}}
\hfill
\subfigure[Hypertension training set with cosine distance. \label{hypertension-train-c}]{\includegraphics[scale=0.14]{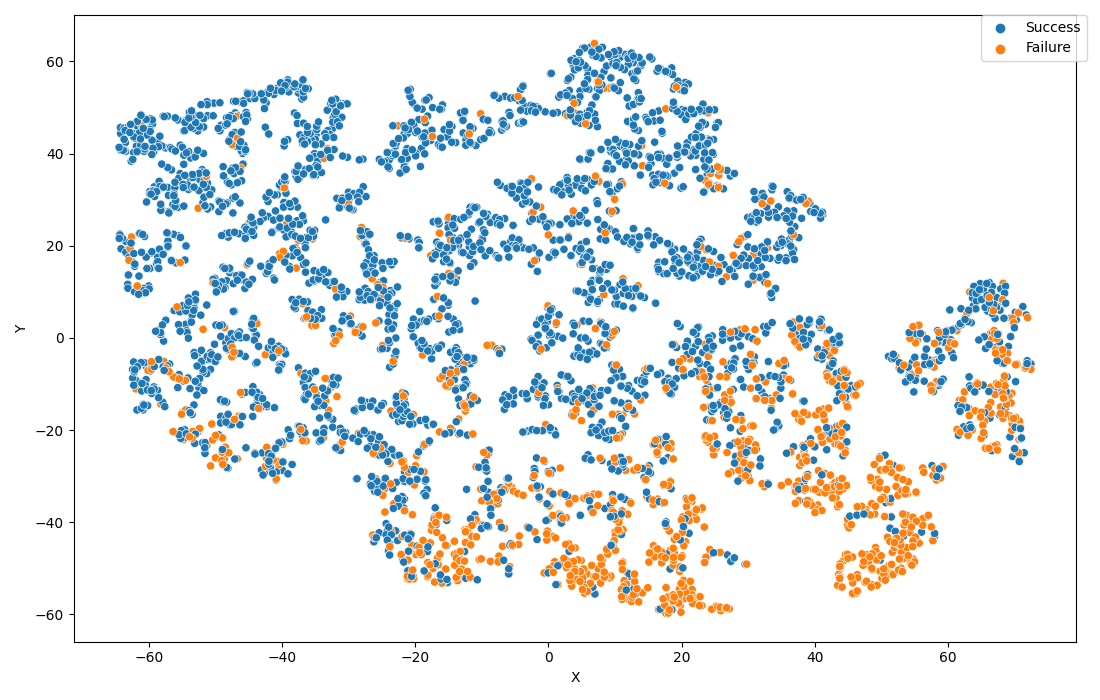}}
\hfill
\subfigure[Hypertension testing set with Euclidean distance. \label{hypertension-test-e}]{\includegraphics[scale=0.14]{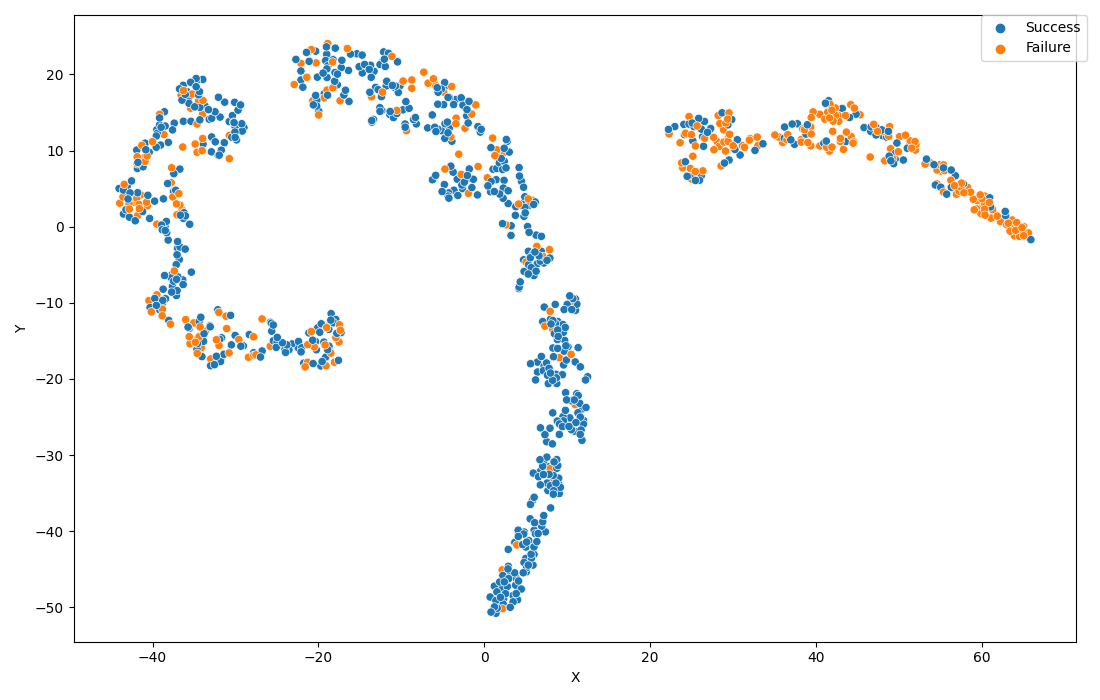}}
\hfill
\subfigure[Hypertension testing set with cosine distance. \label{hypertension-test-c}]{\includegraphics[scale=0.14]{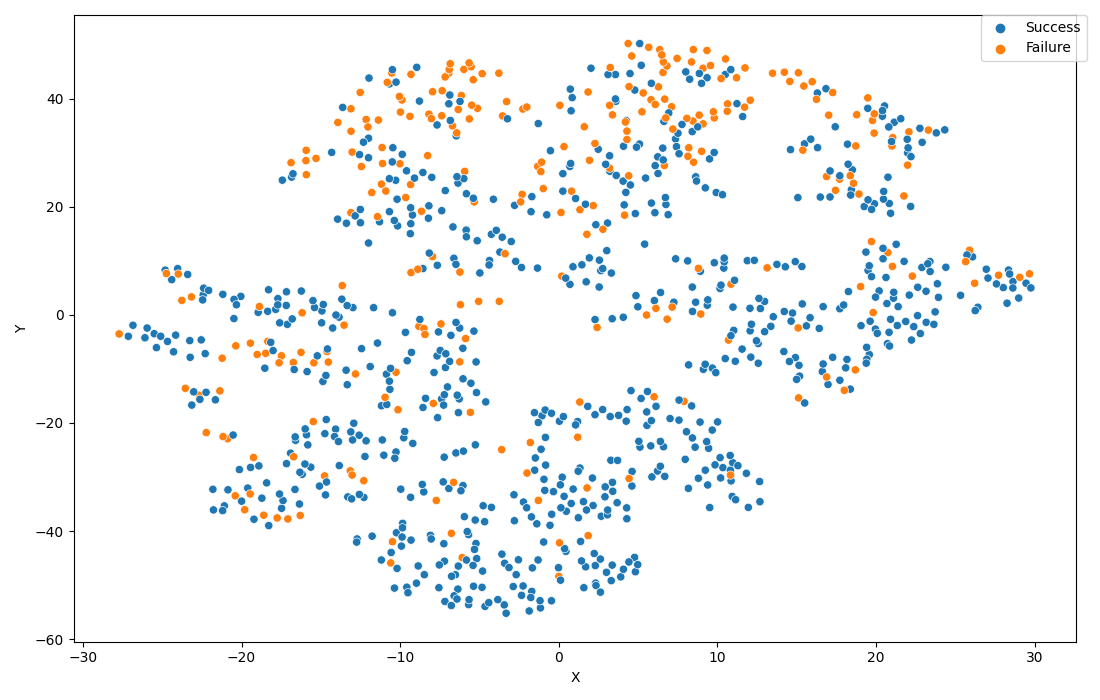}}
\centering
\hfill
\subfigure[Hyperlipidemia training set with Euclidean distance. \label{hyperlipidemia-train-e}]{\includegraphics[scale=0.14]{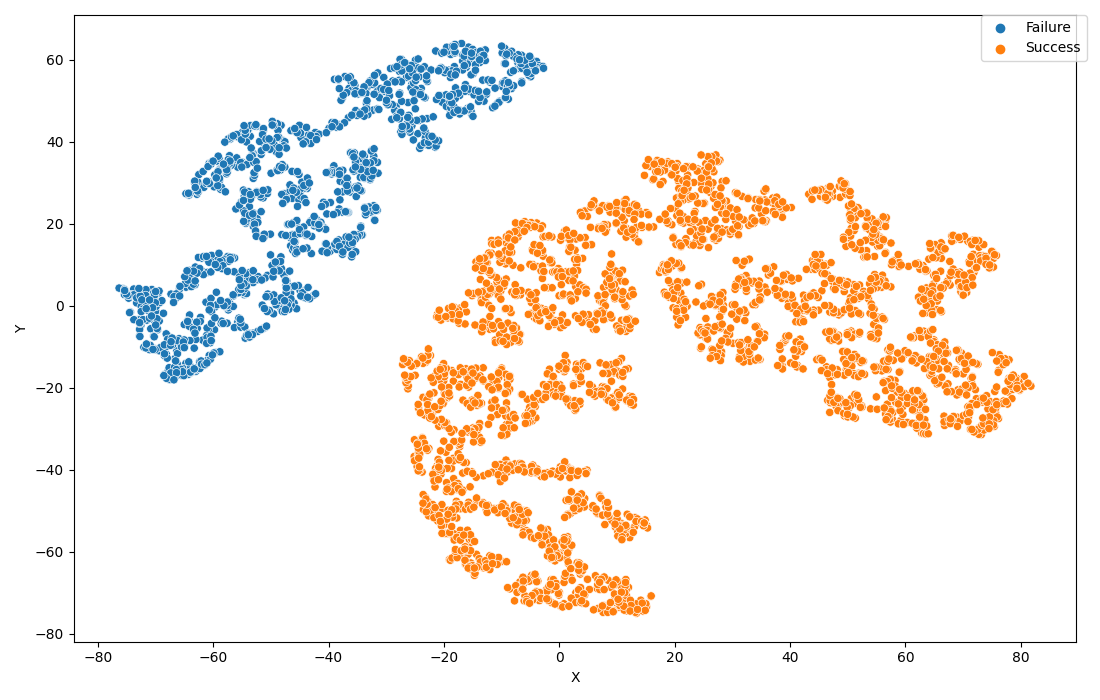}}
\hfill
\subfigure[Hyperlipidemia training set with cosine distance. \label{hyperlipidemia-train-c}]{\includegraphics[scale=0.14]{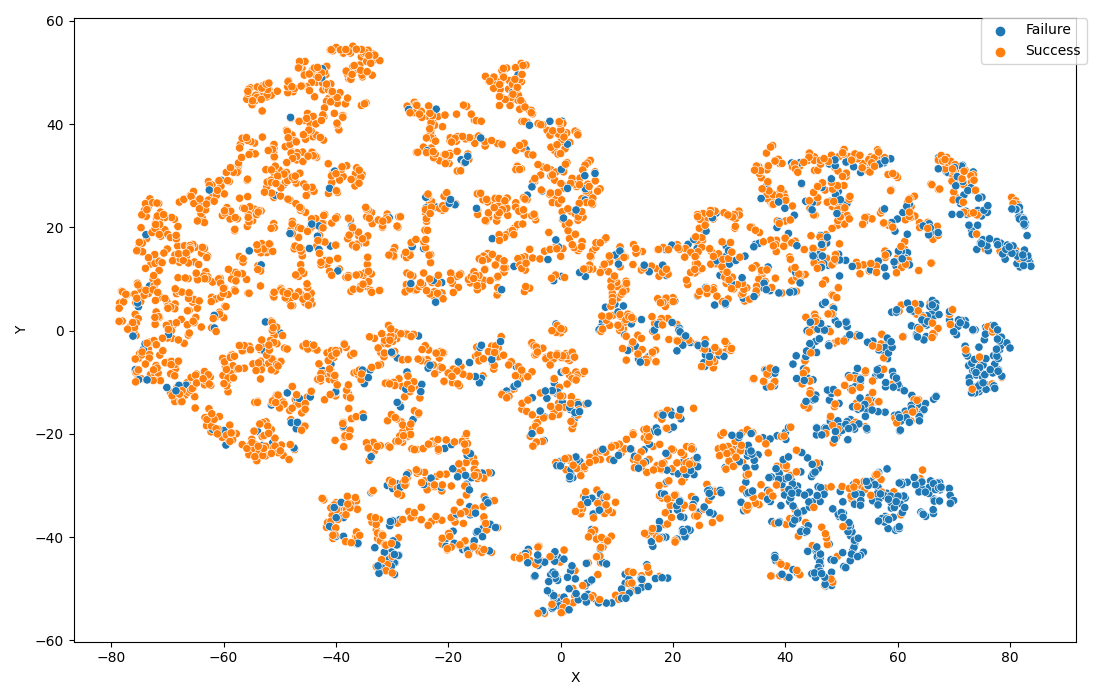}}
\hfill
\subfigure[Hyperlipidemia testing set with Euclidean distance. \label{hyperlipidemia-test-e}]{\includegraphics[scale=0.14]{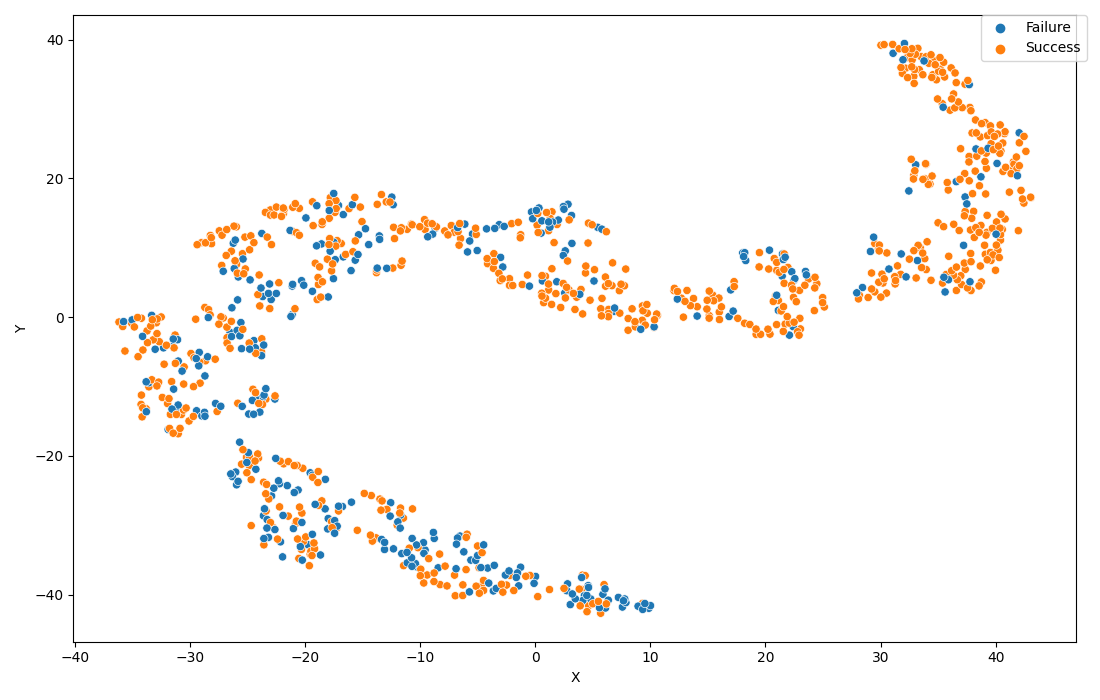}}
\hfill
\subfigure[Hyperlipidemia testing set with cosine distance. \label{hyperlipidemia-test-c}]{\includegraphics[scale=0.14]{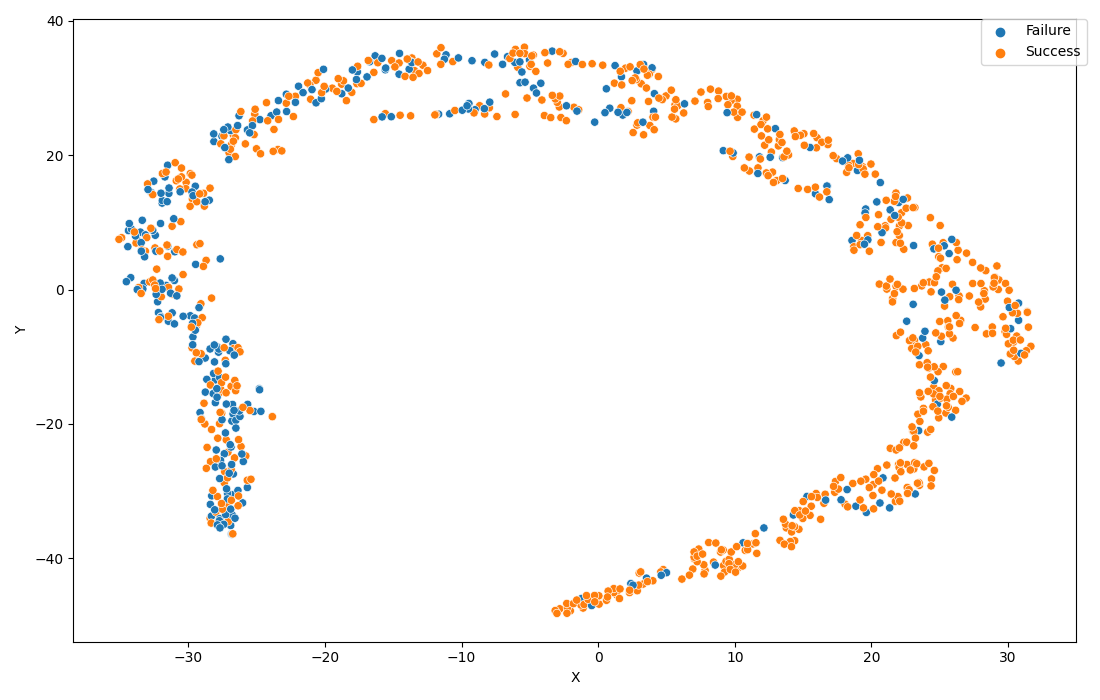}}
\centering
\hfill
\subfigure[Diabetes training set with Euclidean distance. \label{diabetes-train-e}]{\includegraphics[scale=0.14]{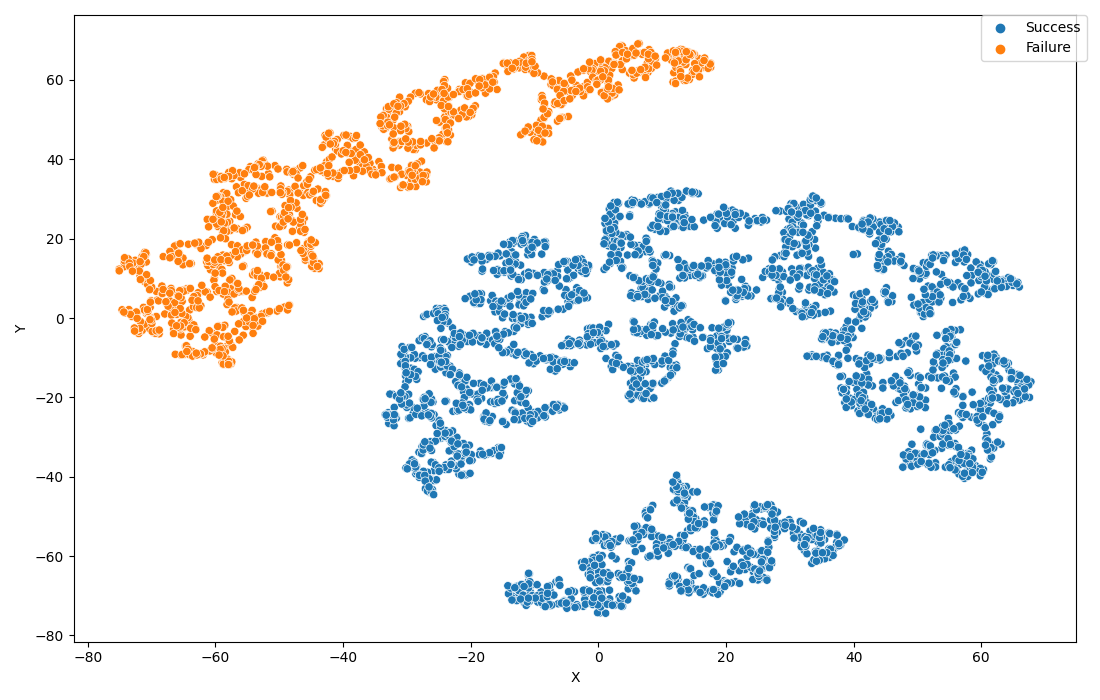}}
\hfill
\subfigure[Diabetes training set with cosine distance. \label{diabetes-train-c}]{\includegraphics[scale=0.14]{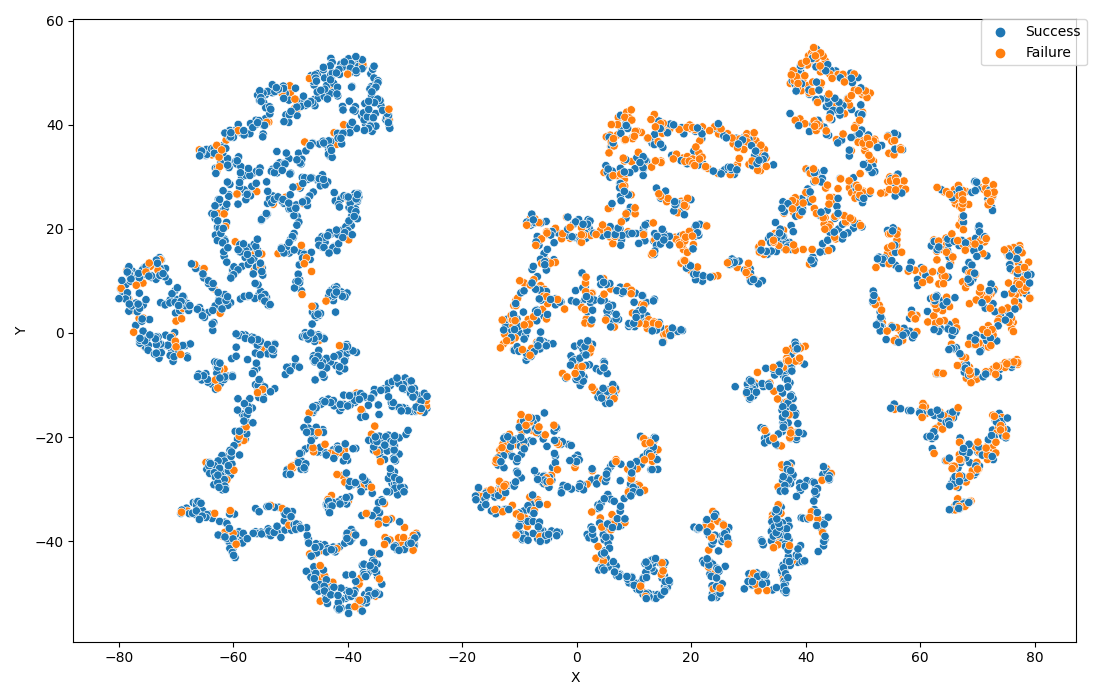}}
\hfill
\subfigure[Diabetes testing set with Euclidean distance. \label{diabetes-test-e}]{\includegraphics[scale=0.14]{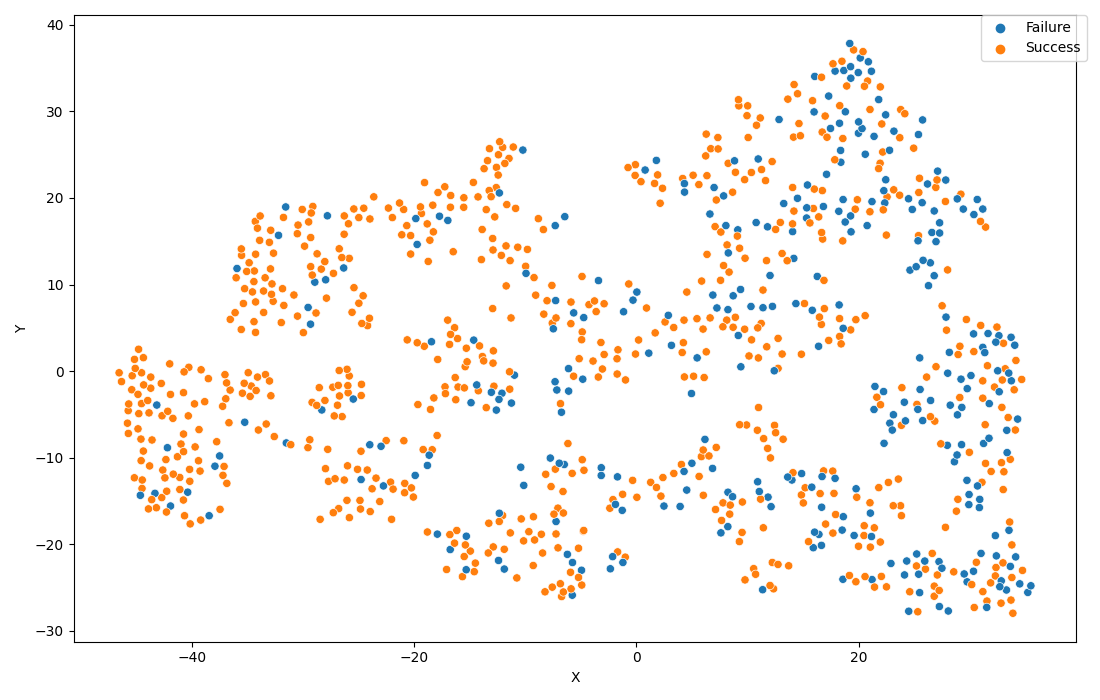}}
\hfill
\subfigure[Diabetes testing set with cosine distance. \label{diabetes-test-c}]{\includegraphics[scale=0.14]{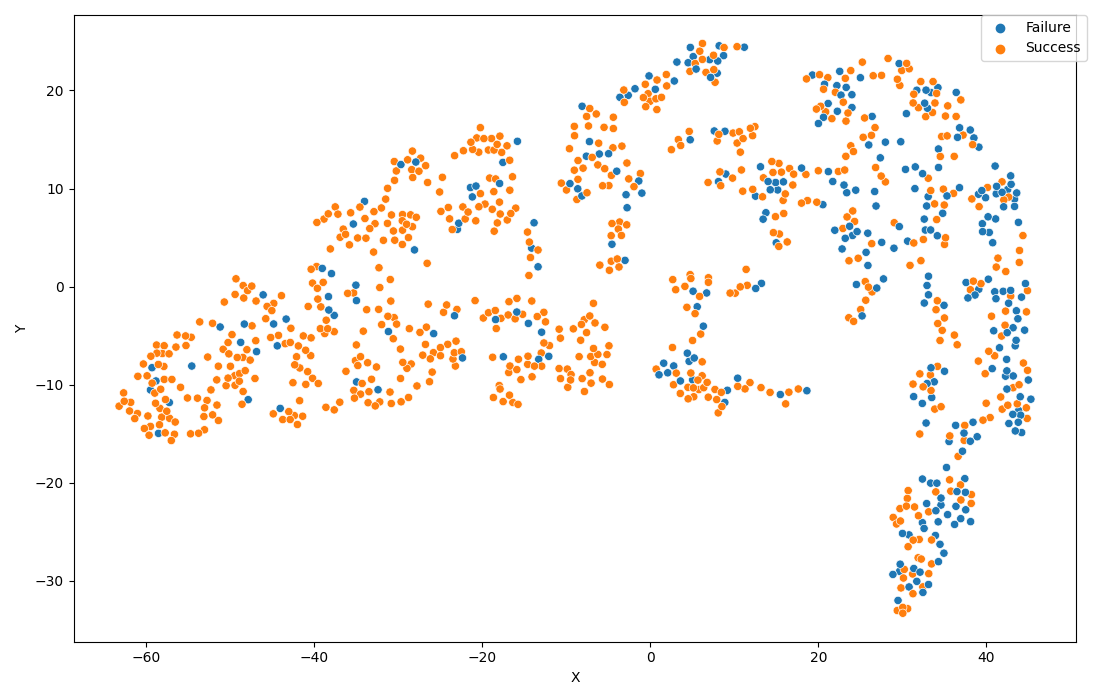}}
\caption{2D visualization of kernel embedding by tSNE with different distance metrics under long-term chronic diseases. We fit our unified framework on training set and evaluate out-of-sample embedding ability on testing set. Blue dot denotes failure case, and orange dot denotes success case.}
\label{long-term-embedding}
\end{figure}

We further compare the embedding space induced by Euclidean and cosine distance measure under their original data-imbalance condition (refer to Table ~\ref{statistics}). We apply t-SNE ~\cite{tsne} algorithm to perform 2D visualization for all diseases under balance and imbalance conditions on the learned kernel embedding from ${F_{emb}}$ in ~\ref{unified-model}. We randomly divide each disease into training and testing in an 80:20 ratio and employ the same training procedure described in ~\ref{evaluation}. Figure ~\ref{short-term-embedding} and ~\ref{long-term-embedding} show the kernel embedding on the training and testing sets with the Euclidean and cosine distance measures under short-term and long-term chronic diseases. In each Figure, the ability to handle out-of-sample (unseen) embedding is depicted with testing set in the sub-title. We can see both Euclidean and cosine distance perfectly separate two outcome groups on training set in short-term disease from Figures ~\ref{ut-train-e},~\ref{ut-train-c},~\ref{aom-train-e},~\ref{aom-train-c},~\ref{bp-train-e},~\ref{bp-train-c},~\ref{ac-train-e}, and ~\ref{ac-train-c}. Their analogous out-of-sample embedding ability in Figures ~\ref{ut-test-e},~\ref{ut-test-c},~\ref{aom-test-e},~\ref{aom-test-c},~\ref{bp-test-e},~\ref{bp-test-c},~\ref{ac-test-e}, and ~\ref{ac-test-c} demonstrate the similar effect between the two distance measures under short-term disease, correlating the evaluation results in tables ~\ref{table-ut},~\ref{table-aom},~\ref{table-bp}, and ~\ref{table-ac}. It also suggests that Euclidean distance is not sufficiently sensitive to highly variant features under long-term disease progressions in chronic disease. As shown in Figures ~\ref{hypertension-test-e},~\ref{hyperlipidemia-test-e}, and ~\ref{diabetes-test-e}, the out-of-sample embedding ability of Euclidean distance under-performs as compared to its cosine distance counterpart in Figures ~\ref{hypertension-test-c},~\ref{hyperlipidemia-test-c}, and ~\ref{diabetes-test-c}, although it can perfectly fit the training set in Figures ~\ref{hypertension-train-e},~\ref{hyperlipidemia-train-e}, and ~\ref{diabetes-train-e}. Cosine distance exhibits certain regularization ability to avoid model overfitting on the training set in chronic disease (refer to Figures ~\ref{hypertension-train-c},~\ref{hyperlipidemia-train-c}, and ~\ref{diabetes-train-c}), resulting in better performance on out-of-sample embedding, which also coincides the evaluation results in Tables ~\ref{table-hypertension},~\ref{table-hyperlipidemia}, and ~\ref{table-diabetes}. It further shows that cosine distance benefits from more complex data representations and Euclidean distance can only be applied to local problems with low data variation, which may not be sufficient to express complex feature characteristics.

\section{Conclusion}
Distance metrics and their nonlinear variant play a crucial role in machine learning based real-world problem solving. We demonstrated how Euclidean and cosine distance measures differ not only theoretically but also in real-world medical application, namely, outcome prediction of drug prescription. Euclidean distance exhibits favorable properties in the local geometry problem. To this regard, Euclidean distance can be applied under short-term disease with low-variation outcome observation. Moreover, when presenting to highly variant chronic disease, it is preferable to use cosine distance. These different geometric properties lead to different submanifolds in the original embedded space, and hence, to different optimizing nonlinear kernel embedding frameworks. We first established the geometric properties that we needed in these frameworks. From these properties interpreted their differences in certain perspectives. Our evaluation on real-world, large-scale electronic health records and embedding space visualization empirically validated our approach.

\bigskip
\noindent{\bf Acknowledgments}
\noindent  The author is grateful to the reviewers for useful suggestions which improved the contents of this paper. 
The first author is partially supported by an NSF grant DMS-1408839 and a McDevitt Endowment Fund at Georgetown University.


\begin{thebibliography}{99}%in appearance order.

\bibitem{drug-1} J. K. Aronson, "Medication errors: what they are, how they happen, and how to avoid them", QJM: An International Journal of Medicine 102.8 (2009): 513-521.

\bibitem{CC1} O. Calin, D.C. Chang, "Sub-Riemannian Geometry, General Theory and Examples", Encyclopedia of Mathematics and Its
Applications,  Cambridge University Press, 126, 2009.

\bibitem{CC2} O. Calin, D.C. Chang, "Geometric Mechanics on Riemannian Manifolds: Applications to Partial Differential Equations", 
Applied and Numerical Analysis, 29, Birkh\"auser, Boston, Massachusetts, 2004.

\bibitem{CCFI} O. Calin, D.C. Chang, K. Furutani, C. Iwasaki, "Heat Kernels for Elliptic and Sub-elliptic Operators: Methods and Techniques", Applied and Numerical Analysis, {\bf 47},  Birkh\"auser, Boston, Massachusetts, 2010.

\bibitem{CCT} O. Carath\'eodory, "Untersuchungen \"uber die Grundlagen der Thermodynamik", Math. Anal., {\bf 67}, 93-161, (1909).

\bibitem{JNCA} D.C. Chang, O. Frieder and H.R. Yao. "On bochner's theorem and its application to graph kernels", Journal of Nonlinear and Convex Analysis 19.12 (2018): 2135-2151.

\bibitem{CLY} D.C. Chang, K.P. Lin and Stephen S.T. Yau: "Schr\"odinger equation with quartic potential and nonlinear
filtering problem", Proceedings of the 48th IEEE Conference on Decision and Control, Shanghai, P.R. China, December 16-18, 8089-8094, (2009).

\bibitem{CL1} D.C. Chang and Y. Li: "Heat Kernels for a family of Grushin operators", Methods and Applications of Analysis, {\bf 21}, $\#${\bf 3}, 291-312, (2014). 

\bibitem{CMV} D.C. Chang, I. Markina, and A. Vasil\'ev: "Geodesics and distance respecting Hopf fibration of $n$-spheres", Journal of Geometry $\&$ Physics, {\bf 61}, $\#${\bf  6}, 986-1000, (2011).

\bibitem{CHOW} W.L. Chow, "Uber Systeme van Linearen partiellen Differentialgleichungen erster Ordnung", Math. Ann., {117}, 98-105, (1939).

\bibitem{k-means} J. Han, M. Kamber, and J. Pei, "Data mining concepts and techniques third edition", The Morgan Kaufmann Series in Data Management Systems 5.4 (2011): 83-124.

\bibitem{dist-sub-kernel} B. Haasdonk, and C. Bahlmann, "Learning with distance substitution kernels", Joint pattern recognition symposium. Springer, Berlin, Heidelberg, (2004).

\bibitem{computer-vision-related} J. Hu, J. Lu, and Y.P. Tan, "Discriminative deep metric learning for face verification in the wild", Proceedings of the IEEE conference on computer vision and pattern recognition (2014).

\bibitem{cf-loss} R. Hadsell, S. Chopra, and Y. LeCun, "Dimensionality reduction by learning an invariant mapping", 2006 IEEE Computer Society Conference on Computer Vision and Pattern Recognition (2016). Vol. 2. IEEE, 2006.

\bibitem{deep-kl1} M. Jiu, and H. Sahbi, "Nonlinear deep kernel learning for image annotation", IEEE Transactions on Image Processing 26.4 (2017): 1820-1832.

\bibitem{drug-2} H.A. Jackson, J. Cashy, O. Frieder, and A.J. Schaeffer, "Data mining derived treatment algorithms from the electronic medical record improve theoretical empirical therapy for outpatient urinary tract infections", The Journal of urology 186.6 (2011): 2257-2262. 

\bibitem{metric-learning-survey} B. Kulis, "Metric learning: A survey", Foundations and trends in machine learning 5.4 (2012): 287-364.

\bibitem{gk-survey} N.M. Kriege, F.D. Johansson, and C. Morris, "A survey on graph kernels", Applied Network Science 5.1 (2020): 1-42.

\bibitem{adam} D.P. Kingma, and J. Ba, "Adam: A Method for Stochastic Optimization", 2015 International Conference on Learning Representations (2015).

\bibitem{p2v} Q. Le, and T. Mikolov, "Distributed representations of sentences and documents", International conference on machine learning. PMLR (2014).

\bibitem{deep-metric-learning} X. Liu, B.V.K. Vijaya Kumar, J. You, and P. Jia, "Adaptive deep metric learning for identity-aware facial expression recognition", Proceedings of the IEEE Conference on Computer Vision and Pattern Recognition Workshops (2017).

\bibitem{ruocco} A. S. Ruocco, and O. Frieder, "Clustering and classification of large document bases in a parallel environment", Journal of the American Society for Information Science. (1997).

\bibitem{text-mining-survey} A. Singhal, "Modern information retrieval: A brief overview", IEEE Data Eng. Bull. 24.4 (2001): 35-43.

\bibitem{wl-kernel} N. Shervashidze, P. Schweitzer, E.J. Van Leeuwen, K. Mehlhorn, and K.M. Borgwardt, "Weisfeiler-lehman graph kernels", Journal of Machine Learning Research 12.9 (2011).

\bibitem{vh-kernel} M. Sugiyama and K.M. Borgwardt, "Halting in random walk kernels", Proceedings of the 28th International Conference on Neural Information Processing Systems-Volume 1. (2015).

\bibitem{Tele} C. Teleman: "Asupra Sistemelor Mecanice Neonolome",  Anal. Univ. `{\it `C.I. Purhon}",  Bucuresti, Seria St. Naturii, {\bf 13}, 45-52, (1957).

\bibitem{tsne} L. Van der Maaten, and G. Hinton, "Visualizing data using t-SNE", Journal of machine learning research 9.11 (2008).

\bibitem{deep-kl2} A.G. Wilson, Z. Hu, R. Salakhutdinov, and E.P. Xing, "Deep kernel learning", Artificial intelligence and statistics. PMLR (2016).

\bibitem{bhi-2019} H.R. Yao, D.C. Chang, O. Frieder, W. Huang, and T.S. Lee, "Graph Kernel prediction of drug prescription", 2019 IEEE EMBS International Conference on Biomedical \& Health Informatics (BHI). IEEE (2019).

\bibitem{bcb-2019} H.R. Yao, D.C. Chang, O. Frieder, W. Huang, and T.S. Lee, "Multiple Graph Kernel Fusion Prediction of Drug Prescription", Proceedings of the 10th ACM International Conference on Bioinformatics, Computational Biology and Health Informatics (2019).

\bibitem{bcb-2020} H.R. Yao, D.C. Chang, O. Frieder, W. Huang, I.C. Liang, and C.F. Hung, "Cross-Global Attention Graph Kernel Network Prediction of Drug Prescription", Proceedings of the 11th ACM International Conference on Bioinformatics, Computational Biology and Health Informatics (2020).


\end{thebibliography}
\end{document}